\pdfoutput=1   
\documentclass[10pt,twocolumn,letterpaper]{article}
\usepackage[pagenumbers]{cvpr}     
\usepackage{research-markers}      
\usepackage{colortbl}
\usepackage{array}
\usepackage{algorithm}
\usepackage{algpseudocode}
\usepackage{placeins}
\usepackage{stfloats}
\usepackage{microtype}
\newcolumntype{N}[1]{>{\raggedleft\arraybackslash}p{#1}}
\newcolumntype{C}[1]{>{\centering\arraybackslash}p{#1}}
\newcommand{\tb}[1]{{\fontseries{b}\selectfont #1}}
\definecolor{cvprblue}{rgb}{0.21,0.49,0.74}
\usepackage[breaklinks,colorlinks,allcolors=cvprblue]{hyperref}
\title{Learning to Edit IFC Building Models: Verifiable Checkers as the Training Signal for Dependency-Aware Editing Agents}

\author{Jiaxin Zhang\textsuperscript{1}\quad
Yao Luo\textsuperscript{1}\quad
Yuze Dan\textsuperscript{2,3}\thanks{Corresponding author: \texttt{danyuze@cqu.edu.cn}}\quad
Yunqin Li\textsuperscript{1}\quad
Bowen Wang\textsuperscript{4}\\[4pt]
{\small \textsuperscript{1}Architecture and Design College, Nanchang University, Nanchang, China}\\
{\small \textsuperscript{2}School of Architecture and Urban Planning, Chongqing University, Chongqing, China}\\
{\small \textsuperscript{3}Key Laboratory of New Technology for Construction of Cities in Mountain Area (Ministry of Education), Chongqing University, China}\\
{\small \textsuperscript{4}Department of Intelligent Media, SANKEN, The University of Osaka, Osaka, Japan}
}

\begin{document}
\maketitle
\begin{abstract}
Editing a building information model in the Industry Foundation Classes (IFC) schema requires propagating changes to dependent elements while preserving unrelated entities. Existing language-model agents for IFC editing use prompted models without learning from edit outcomes. Target-file comparisons do not specify which alternative edits are allowed or which unrelated entities must remain unchanged. We present IfcGym, a CodeAct environment in which a language model edits IFC through executable code and four checkers verify validity relative to the input model, intent, dependency and preservation against a task specification. Their conjunction is strict success. IfcEdit distils verified teacher episodes into a Qwen3.5-9B student through supervised fine-tuning (SFT), then applies Group Relative Policy Optimization (GRPO) with strict success as the reward. \claimref{C9012}\evref{E019} On IfcEditBench, 1,990 editing tasks on 13 held-out real projects, the final student reaches 86.78\% strict success, from 79.15\% after SFT. The paired confidence interval for the difference from its teacher includes zero; \claimref{C9013}\evref{E024} the same recipe on a 27B model reaches 91.76\% and exceeds the teacher. \claimref{C9010}\evref{E019} On the public benchmarks BIM-Edit and BIBIMBAP the final student scores 55.19 and 54.50. On IfcEditBench, the students outperform all prompted models except the teacher under matched tool access; both evaluated frameworks also score lower. \claimref{C9012}\evref{E019}\evref{E022} Rewarding all four checkers yields 4.83 points more strict success than rewarding intent and validity alone.
\end{abstract}

\section{Introduction}
\label{sec:introduction}

Building information modelling (BIM) uses digital representations of built assets to support information management and collaboration \cite{borrmann2018bim,wang2024improving}. The Industry Foundation Classes (IFC) standard defines an open, vendor-neutral data schema for representing and exchanging this information \cite{isoifc2024}; an IFC model represents a building as typed, interrelated entity instances. Design revisions, code-compliance checking and repair \cite{eastman2009rule}, and coordination across disciplines require changes to existing models. Such changes rarely affect an isolated element. Moving a wall may require moving its openings, doors and windows and updating the boundaries of adjacent spaces; deleting a door may also require removing its opening. We study \emph{dependency-aware editing}: carrying out a natural-language instruction, propagating the change to its specified dependents, and preserving unrelated entities.

Dependency-aware editing requires reasoning over the model's entity graph. Several edit sequences can satisfy an instruction, while a geometrically plausible result can still leave an opening detached, a space boundary inconsistent or an unrelated entity altered. Existing approaches encode building dependencies in a manually specified topology graph \cite{alterations2026} or learn to resolve clashes through a small set of rigid moves on one model \cite{bimrl2025}. On an unseen building, the agent must identify which entities an instruction affects and how to edit them while preserving their relationships.

Large language models (LLMs) support BIM question answering and retrieval \cite{zheng2023bimgpt,saka2024gpt,qwenbim2026}, and can access a model through code. CodeAct \cite{codeact2024} and IFC tool servers \cite{mcp4ifc2025,ifcagent2026} provide interfaces for inspecting and editing a model. These IFC agents rely on prompted models and do not train on their edit outcomes. The benchmarks BIM-Edit \cite{bimedit2026} and BIBIMBAP \cite{bibimbap2026} expose failures involving dependent elements and integrity preservation. Learning from these failures requires a success criterion that checks the requested modification, its dependent changes and the entities that must remain unchanged. A single target file does not directly specify this distinction or the range of acceptable solutions.

We make these editing requirements an executable training objective. IfcGym represents each task by a \emph{specification} of its targets, allowed changes, intended modification and dependency rules. Four checkers evaluate validity, intent, dependency and preservation; their conjunction defines \emph{strict success}. The same criterion selects successful teacher demonstrations, rewards the agent during reinforcement learning and evaluates its final edits. Different edit sequences can pass when they satisfy the specification; no target file is needed.

\begin{figure*}[!t]
 \centering
 \includegraphics[width=\linewidth]{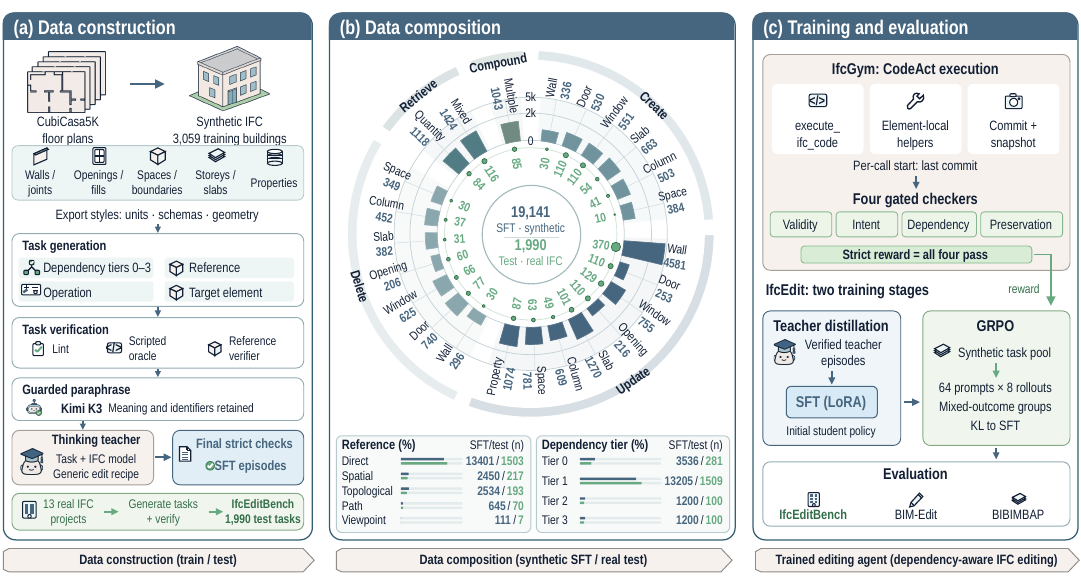}
 \caption{Data construction and learning in IfcGym. (a) Synthetic buildings supply verified teacher demonstrations; real IFC projects supply IfcEditBench tasks. (b) Bar heights encode square-root SFT counts; green circle areas encode test counts. Outer blue and inner green numbers give counts; lower panels show within-split reference/tier proportions and counts. (c) Teacher distillation, strict-reward GRPO and benchmark evaluation.}
 \label{fig:overview}
\end{figure*}

IfcEdit uses this supervision to train an open language model for dependency-aware editing. It first distils verified teacher episodes into Qwen3.5-9B through supervised fine-tuning (SFT) \cite{hinton2015kd,star2022,studentdistill2025}, then improves the policy through Group Relative Policy Optimization (GRPO) with strict success as the reward \cite{deepseekmath2024,retool2025,toolcad2026,dapo2025}. \claimref{C9004}\evref{E002}\evref{E003}\evref{E004} The SFT corpus contains 19,141 verified episodes on 3,059 synthetic buildings with geometry and naming conventions drawn from real exports. \claimref{C9012}\evref{E019} Evaluation tests transfer to real buildings through IfcEditBench, comprising 1,990 tasks on 13 projects excluded from training, and to the public BIM-Edit and BIBIMBAP benchmarks. Figure~\ref{fig:overview} connects synthetic training and real-project test construction in (a), the resulting data composition in (b), and agent training and evaluation in (c).

\claimref{C9004}\evref{E002}\evref{E003}\evref{E004} We collect verified demonstrations across synthetic buildings with varied layouts and export conventions. To distinguish what the students learn from what the tools provide, we give prompted and trained models the same editing operations. We then compare each student before training, after SFT and after GRPO to measure the gains from the two stages.

\claimref{C9012}\evref{E019}\evref{E021} On IfcEditBench, Qwen3.5-9B improves from 16.08\% strict success when prompted to 79.15\% after SFT and 86.78\% after GRPO; the best prompted model other than the teacher with the same CodeAct tools reaches 67.99\%, and the teacher 88.04\%. \claimref{C9014}\evref{E019}\evref{E021}\evref{E022}\evref{E024} The largest gaps occur on edits that must propagate to dependent elements, which the other prompted models and evaluated IFC frameworks often fail to complete. \claimref{C9012}\evref{E019} The paired gain from GRPO is 7.75 points, comparable in magnitude to the 9.15-point gain from applying the same SFT recipe to the larger 27B model. \claimref{C9013}\evref{E024} Applying GRPO to that larger student raises strict success further, to 91.76\%.

\claimref{C9010}\evref{E019} Improvements also transfer to the public benchmarks: the final 9B student scores 55.19 on BIM-Edit and 54.50 on BIBIMBAP, improving over its SFT start on both. \claimref{C9012}\evref{E019}\evref{E022} In the reward ablation, the four-gate reward improves strict success by 4.83 points over rewarding intent and validity alone.

The contributions are:
\begin{enumerate}
 \item IfcGym turns requested edits, dependency propagation and preservation into executable supervision, providing a common criterion for constructing verified training data, rewarding edits and evaluating outcomes.
 \item \claimref{C9012}\evref{E019}\claimref{C9013}\evref{E024} IfcEdit combines teacher distillation and GRPO to train open models for dependency-aware editing, with gains beyond SFT at both 9B and 27B scales.
 \item \claimref{C9012}\evref{E019}\evref{E021} Evaluation on IfcEditBench and two public benchmarks measures transfer from synthetic buildings to real projects and compares the students with their teacher and other prompted models under matched tool access. Paired uncertainty estimates and reward-term and data-scale ablations quantify these differences.
\end{enumerate}

\section{Related work}
\label{sec:related}

\subsection{Language-model agents and benchmarks for IFC models}
\label{sec:rw-ifc}

Language-model work in BIM includes question answering and information retrieval over IFC \cite{zheng2023bimgpt,ifcbench2026}, domain-adapted models and benchmarks \cite{qwenbim2026}, and generation from text or sketches \cite{text2bim2025}. Tool-based agents also give prompted models direct access to IFC files. MCP4IFC \cite{mcp4ifc2025} exposes a tool server for scene queries and attribute edits. \claimref{C009}\evref{E013} Its best configuration answers 83.1\% of 65 scene-querying questions; edits requiring geometry, spatial context or element dependencies were not tested. IFC-Agent \cite{ifcagent2026} adds schema-guided reasoning, isolation and versioning around IfcOpenShell, with evaluations centred on retrieval and generation. KIR \cite{kir2026} proposes a typed intermediate representation and dependency-closure diffs for agent-authored buildings, without an LLM experiment. These systems do not train the model on the outcome of its edits.

BIM-Edit \cite{bimedit2026} and BIBIMBAP \cite{bibimbap2026} evaluate editing more directly. \claimref{C005}\evref{E009} BIM-Edit compares geometry, semantics and topology with one target file per task and reports low strict completion across models. BIBIMBAP covers create, read, update and delete (CRUD) operations with executable per-task tests. \claimref{C006}\evref{E010} Its best model reaches 50.2\% overall, with 81.1\% for Create and 18.9\% for Delete; integrity preservation is the hardest criterion across models. Our setting makes the allowed changes and dependency rules explicit in a task specification. Edits pass when they satisfy those constraints and preserve entities outside the allowed set. We adopt BIM-Edit's CodeAct protocol and use both benchmarks as external tests (Section~\ref{sec:external}); the training reward comes from our entity-graph checkers (Section~\ref{sec:checkers}).

\subsection{Modifying building and CAD models under constraints}
\label{sec:rw-edit}

Wu et al.\ \cite{alterations2026} formalise human alteration strategies with an LLM, apply them through a rule-coded topology graph, and re-check compliance with a commercial model checker. \claimref{C007}\evref{E011} Their adjustment mechanism reduces the sampling cost of feasible alterations, with manually coded dependencies and evaluation on one building. IFCDecoupling \cite{ifcdecoupling2026} addresses safe modification of shared references on the IFC graph. For clash resolution, \claimref{C008}\evref{E012} a PPO agent in a Gym wrapper uses eight rigid move actions and a scalar clash reward, requiring thousands to tens of thousands of steps on a single IFC house \cite{bimrl2025}. Text2BIM \cite{text2bim2025} repairs models during generation, and defect-repair systems \cite{bimdefects2026} produce textual suggestions.

Language models also generate and edit parametric CAD through sketch-and-extrude programs \cite{deepcad2021}, text-to-CAD generation \cite{text2cad2024} and CAD-kernel tools \cite{cadassistant2024}. LLM4CAD-Editor \cite{llm4cadeditor2026} and B-repLer \cite{brepler2025} evaluate language-guided edits through parsing or validity and geometric fidelity. \claimref{C011}\evref{E015} \claimref{C012}\evref{E016} Both report high parsing or validity rates alongside lower fidelity for more functional edits. These metrics do not directly measure preservation of unmentioned entities. City Editing \cite{cityediting2026} evaluates prompted agents under dependencies and constraints in urban geospatial models. ALAS \cite{alas2025} limits repair to a verified neighbourhood and rolls back failures, while trace-guided CAD repair \cite{tracecad2026} bounds edits with a dependency graph and uses a vision-language model to judge preservation from renders. IfcGym checks the edited IFC entities directly after transactional execution. IfcEdit learns how to propagate changes from instructions, with evaluation on held-out projects.

\subsection{Training tool agents with verifiable rewards}
\label{sec:rw-rlvr}

IfcEdit builds on reasoning interleaved with tool calls \cite{react2023,wang2025taming}, executable code actions \cite{codeact2024}, and interfaces and benchmarks for software and service agents \cite{swebench2024}. Reinforcement learning with verifiable rewards \cite{deepseekr12025,dapo2025} has been applied to strategic tool use \cite{retool2025}, while trajectory distillation transfers agent behaviour to smaller models \cite{studentdistill2025}. Ma et al.\ \cite{toolrl2026} find that the gains of tool-use reinforcement learning can arise from reasoning rather than from tool use itself.

In text-to-CAD, ToolCAD \cite{toolcad2026} combines SFT and GRPO with a learned outcome reward. \claimref{C013}\evref{E017} Its 7B and 8B models achieve 61.8--63.9\% success, within a few points of prompted GPT-4o. Embodied CAD \cite{embodiedcad2026} reports little improvement from GRPO over SFT for parametric assembly modelling. Ishigaki-IDS \cite{ishigaki2026} applies validator-gated GRPO to specification drafting on Ishigaki-Bench \cite{ishigakibench2026}. \claimref{C010}\evref{E014} Its 8B model exceeds the frontier baselines in validator pass rate while a content metric remains flat. In floor-plan generation \cite{floorplanrlvr2026}, \claimref{C014}\evref{E018} verifiable rewards improve constraint adherence, but connectivity-only rewards permit reward hacking \cite{rewardgaming2022}. We combine four specification-based checkers in the reward and test the contribution of dependency and preservation checks against an intent-and-validity reward (Section~\ref{sec:ablations}).

\section{Method}
\label{sec:method}
\suppressfloats[t]

\begin{table}[t]
\caption{The helper package \texttt{h}: 28 helpers in three groups.}
\label{tab:tools}
\footnotesize
\setlength{\tabcolsep}{3pt}
\renewcommand{\arraystretch}{1.15}
\begin{tabular*}{\linewidth}{@{}>{\raggedright\arraybackslash}p{0.2\linewidth}>{\raggedright\arraybackslash}p{\dimexpr0.8\linewidth-2\tabcolsep\relax}@{}}
\toprule
\textbf{Group} & \textbf{Helpers} \\
\midrule
Inspection (read-only) & \texttt{inspect}, \texttt{find}, \texttt{unit\_scale}, \texttt{world\_matrix}, \texttt{wall\_axis\_world}, \texttt{profile\_world\_polygon}, \texttt{wall\_length}, \texttt{space\_area}, \texttt{check\_model} \\
Element-local edits & \texttt{translate}, \texttt{move\_wall\_end}, \texttt{stretch\_wall\_end}, \texttt{shift\_vertices\_on\_plane}, \texttt{shift\_profile\_vertices}, \texttt{shift\_boundary\_curve}, \texttt{set\_extrusion\_depth}, \texttt{set\_wall\_thickness}, \texttt{rotate}, \texttt{set\_property}, \texttt{set\_attribute}, \texttt{delete}, \texttt{connect\_walls}, \texttt{private} \\
Creation & \texttt{add\_wall}, \texttt{add\_slab}, \texttt{add\_column}, \texttt{add\_space}, \texttt{add\_opening\_fill} \\
\bottomrule
\end{tabular*}
\end{table}

We formulate dependency-aware IFC editing as code generation under a verifiable specification (Section~\ref{sec:problem}). IfcGym executes and checks the agent's edits (Section~\ref{sec:ifcgym}), supports task and data construction (Section~\ref{sec:tasks}), and supplies the training signal for IfcEdit (Section~\ref{sec:training}).

\subsection{Problem formulation}
\label{sec:problem}

An editing task is a triple $(\mathcal{M}, q, \sigma)$: an IFC model $\mathcal{M}$, a graph of typed entity instances linked by attribute references and by relationship entities (the subtypes of \texttt{IfcRelationship}); a natural-language instruction $q$; and a specification $\sigma=(T,A,\iota,\Delta)$ that the agent never sees. The specification and the checkers operate on \emph{products} (\texttt{IfcProduct}: elements, openings, spaces, storeys, building and site), keyed by their GlobalId; placement, geometry and property-set entities are folded into the product that owns them, so a change to any of them counts as a change of that product (quantity sets are not compared). $T$ is the set of target products the instruction refers to, $A\supseteq T$ the set of products \emph{allowed} to change, $\iota$ the \emph{intent} (the type of change and its parameters, for example a translation vector, a property value or a new height) and $\Delta=\{(d_j,\rho_j)\}$ the \emph{dependents} that must follow the target, each with the rule $\rho_j$ it obeys: an opening and its fill translate with their host wall and stay hosted, a door's opening is removed with the door, a space face meets the wall that was moved. The generator fixes the editing policy for each task family and states it in both the specification and the instruction. For example, IFC allows an opening to remain after its door is removed; a task can require either removal or retention of that opening. The agent produces a final model $\mathcal{M}'$. Four binary checkers, each a function of $(\mathcal{M},\mathcal{M}',\sigma)$, judge the outcome: validity $v$, intent $i$, dependency $d$ and preservation $p$ (Section~\ref{sec:checkers}). We define \emph{strict success} as their conjunction,
\begin{equation}\label{eq:strict}
  s(\mathcal{M},\mathcal{M}',\sigma)=v\cdot i\cdot d\cdot p ,
\end{equation}
which is the held-out metric and reinforcement reward. Because $\sigma$ specifies constraints rather than a target file, any $\mathcal{M}'$ satisfying them passes, regardless of the API-call sequence.

\subsection{IfcGym}
\label{sec:ifcgym}

IfcGym supports teacher-episode collection, student training and evaluation through a transactional sandbox, element-local helper tools and the four checkers.

\subsubsection{Sandbox}
\label{sec:sandbox}

The agent has a single tool, \texttt{execute\_ifc\_code}, following BIM-Edit's CodeAct interface \cite{bimedit2026}. Inside the code passed to it, the agent calls the helper package \texttt{h} (Table~\ref{tab:tools}) or IfcOpenShell directly, and closes an editing step with \texttt{commit()}; a call without \texttt{commit()} only inspects the model. The tool runs Python in a subprocess with the model loaded through IfcOpenShell \cite{ifcopenshell}. It returns standard output and the value assigned to \texttt{result}; changes persist only when the code calls \texttt{commit()}. Let $\mathcal{M}_k$ denote the model after the $k$-th commit. Each call $c$ uses a fresh namespace and starts from the last committed state:
\begin{equation}\label{eq:commit}
  \mathcal{M}_{k+1}=
  \begin{cases}
    c(\mathcal{M}_k) & \text{if } c \text{ calls } \texttt{commit()},\\
    \mathcal{M}_k & \text{otherwise,}
  \end{cases}
  \qquad \mathcal{M}_0=\mathcal{M}.
\end{equation}
Uncommitted changes are discarded. Each commit writes a numbered snapshot, allowing replay and rollback. The final model is the last commit, $\mathcal{M}'=\mathcal{M}_K$, whether the episode ends with an answer or budget exhaustion. The worker has separate user, mount, process and network namespaces, with access restricted to its working directory, helper package and input model. \claimref{C9004}\evref{E002}\evref{E003}\evref{E004} Episodes allow 20 tool-call rounds and terminate when the model answers without a tool call or exhausts that budget.

\subsubsection{Helper tools}
\label{sec:tools}

The helper package \texttt{h} is documented in the system prompt (Table~\ref{tab:tools}); the full signatures are listed in Appendix~\ref{sec:supp_tools}. Inspection returns an entity's placement chain, representation, openings and fills, wall connections, space boundaries, property sets and shared sub-entities. These structured results reduce the exploratory calls needed to recover relationships from unfamiliar IFC attributes.

Edit helpers are \emph{element-local}: they translate a placement, stretch a wall end, move body vertices or a space-boundary surface, set a property, or delete a product and clean its relationships. Shared sub-entities are copied before modification so that changing a shared point does not unintentionally move another wall. The agent still selects which dependent products to edit, which ends to move and the direction of each change.

\subsubsection{Gated checkers and the strict reward}
\label{sec:checkers}

The checkers read $\sigma$, $\mathcal{M}$ and $\mathcal{M}'$. Each returns a binary verdict; all must pass for strict success.

\paragraph{Validity} The checker parses $\mathcal{M}'$ and runs the IfcOpenShell schema validator (attribute types and cardinalities; EXPRESS where-rules are not evaluated) on $\mathcal{M}$ and $\mathcal{M}'$; an error already reported on $\mathcal{M}$ is excluded, so the gate tests that the edit introduced no new schema error. The body geometry of every product the edit touched must rebuild in the IfcOpenShell geometry kernel, and a cycle in the decomposition graph (\texttt{IfcRelAggregates}, \texttt{IfcRelNests}) fails the gate, because geometry generation does not terminate on such a model. Validity here is limited to these schema, geometry and cycle tests. Full IFC or model-view conformance, structural consistency and regulatory compliance are outside their scope.

\paragraph{Intent} The checker tests the requested change on the targets according to $\iota$: translation within tolerance, an attribute or property value, deletion, a dimension change, creation of the requested class at the specified location, or a retrieval answer in the final message. Leaving a requested edit undone fails this check and is also recorded by an \emph{under-edit} indicator.

\paragraph{Dependency} Each dependent $d_j$ in $\Delta$ must satisfy its rule $\rho_j$ in $\mathcal{M}'$, such as translating with the target, retaining its host, being removed or retained, or meeting or remaining joined to a moved wall. The rules test the resulting model. A product placed relative to the moved wall, as an opening is in many exports, translates with it automatically. The delete helper removes the deleted product's relationships; the agent must separately remove the opening of a deleted door. A space face, a space-boundary surface or a joined wall end has its own geometry and must be edited by the agent; the tiers of Table~\ref{tab:taxonomy} are ordered by how many such explicit dependent edits a task requires.

\paragraph{Preservation} An entity-level diff between $\mathcal{M}$ and $\mathcal{M}'$, in which a product counts as modified when its attributes, property sets, representation or placement change, yields the set $\Theta$ of entities added, removed or modified and the sets $R^{+}$ and $R^{-}$ of relationships added and removed. With $\mathcal{A}=T\cup A\cup\{d_j\}\cup N$, where $N$ holds the new entities of the classes a creation task expects, preservation holds when nothing outside $\mathcal{A}$ changed:
\begin{equation}\label{eq:preserve}
\begin{aligned}
  p=\ &\mathbf{1}\bigl[\Theta\setminus\mathcal{A}=\emptyset\bigr]\\
      &\cdot\mathbf{1}\bigl[\forall (a,b)\in R^{+}\cup R^{-}:\ a\in\mathcal{A}\ \vee\ b\in\mathcal{A}\bigr].
\end{aligned}
\end{equation}
The second factor detects relationship removals between entities unrelated to the task, such as a space boundary or wall connection two rooms away. An entity-only diff misses these changes. Relationships with one end in $\mathcal{A}$ are governed by the listed dependency rules. If an allowed product has no dependency rule, its relationship to the target may change. When a wall that bounds a space is moved at tier one, the space is in $A$ but not in $\Delta$, so removing the space boundary between them passes $p$; at tier two the space is in $\Delta$ with the rule that its face meets the moved wall and its boundary follows, so the removed boundary fails $d$. A connection removed between two walls elsewhere on the storey fails $p$ at every tier.

The reinforcement reward is strict success from Eq.~\eqref{eq:strict}: $r(\tau)=s$ for an episode $\tau$ with final commit $\mathcal{M}'$. Different edit sequences receive the same reward when they satisfy the specification.

\subsection{Tasks and data}
\label{sec:tasks}

\begin{table*}[t]
\caption{Task taxonomy: the dependency tier (what the task's policy requires to follow the target; everything outside the allowed set is protected by the preservation gate), how the instruction names its target, what it asks for, and the target elements. The two lists of each block are independent; rows do not correspond across them.}
\label{tab:taxonomy}
\footnotesize
\renewcommand{\arraystretch}{1.15}
\setlength{\tabcolsep}{4pt}
\newcommand{\taxhead}[2]{\multicolumn{2}{@{}>{\columncolor{black!7}[0pt][0pt]}p{\dimexpr0.47\textwidth+2\tabcolsep\relax}@{}}{\rule[-4pt]{0pt}{17pt}\hspace{4pt}\textbf{#1}\hspace{7pt}{\normalfont\itshape #2}}}
\begin{tabular}{@{}>{\raggedright\arraybackslash\bfseries}p{0.09\textwidth} >{\raggedright\arraybackslash}p{0.38\textwidth} @{}p{\dimexpr0.06\textwidth-4\tabcolsep\relax}@{} >{\raggedright\arraybackslash\bfseries}p{0.125\textwidth} >{\raggedright\arraybackslash}p{0.345\textwidth}@{}}
\arrayrulecolor{black}\toprule[0.7pt]
\taxhead{Dependency tier}{What must follow} & & \taxhead{Reference}{How the target is named} \\[3pt]
Tier 0 & attribute or property of one element; only the target may change &  & Direct & by its GlobalId \\
\addlinespace[2.5pt]
Tier 1 & translate a wall with openings, delete a door, change a wall height, create an element; openings and fills translate with the wall and stay hosted; by the task's policy a deleted door's opening is removed and its host wall stays &  & Spatial & by geometry relative to a named entity: the nearest parallel wall north of a wall, the window on the storey above \\
\addlinespace[2.5pt]
Tier 2 & translate a wall that bounds spaces; as tier 1, and each bounded space's face on the wall side meets the moved wall, its space boundary moving with it &  & Topological & through model relationships: the space bounded by a given set of walls, the wall that hosts a given door \\
\addlinespace[2.5pt]
Tier 3 & translate a wall joined to other walls; as tier 2, and each joined wall stays joined to the moved wall &  & Path & along a route through the building: after entering a space through a given door, the wall on your left \\
\addlinespace[2.5pt]
 &  &  & Viewpoint & as seen from a standpoint: the wall facing you when looking in through a given window \\
\addlinespace[7pt]
\taxhead{Operation}{What the task asks for} & & \taxhead{Target element}{What is targeted} \\[3pt]
Create & a new wall, column, slab, space, door or window (with its opening) &  & Wall, slab, column & \texttt{IfcWall} (including \texttt{IfcWallStandardCase}), \texttt{IfcSlab} and \texttt{IfcColumn}, each with its own body and contained in a storey; a wall also carries openings, joints and space boundaries \\
\addlinespace[2.5pt]
Update & move, resize, rotate, re-host or re-attribute an existing element; its dependents must follow &  & Door, window, opening & an opening is the void cut into a host wall; a door or window is the fill hosted by an opening and moves with its wall \\
\addlinespace[2.5pt]
Delete & remove an element with its relationships; its dependents are removed or kept by rule &  & Space & an \texttt{IfcSpace} aggregated into a storey; space tasks are limited to interior rooms whose space boundaries to their walls are explicit \\
\addlinespace[2.5pt]
Retrieve & answer a question (count, length, area, volume, identifier); the model must stay unchanged &  & Property, quantity & a value in an element's property set, or a quantity derived from an element (length, area, volume, count): the targets of attribute updates and of quantity take-off \\
\addlinespace[2.5pt]
Compound & two of the above in one instruction &  &  & \\
\arrayrulecolor{black}\bottomrule[0.7pt]
\end{tabular}
\end{table*}

For SFT, the data-construction pipeline instantiates a task on a synthetic building, verifies the task, paraphrases its prompt under guards and collects a teacher solution (Algorithm~\ref{alg:task}). Only episodes that pass strict success enter the supervised corpus. The separate test-data branch in Figure~\ref{fig:overview}(a) constructs IfcEditBench tasks on held-out real IFC projects and applies the task-verification gates; the corpus and test split are detailed in Section~\ref{sec:corpus}.

\begin{algorithm}[t]
\caption{One task, from draw to training episode. $\mathcal{T}$ denotes the teacher, $\mathcal{C}$ the four checkers and $s$ strict success. Every gate discards on failure.}
\label{alg:task}
\begin{algorithmic}[1]
\State $(\mathrm{template}, b) \gets$ draw a generator family, a tier and a building
\State $(q_0, \sigma) \gets$ instantiate the template on $\mathcal{M}_b$, truth read from the model
\State \textbf{gate} lint: every identifier in $q_0$ and $\sigma$ resolves, parameters are in range
\State \textbf{gate} oracle: a scripted solution passes $\mathcal{C}$ under $\sigma$ and a wrong one fails
\State \textbf{gate} reference: the entities $q_0$ refers to are unique in $\mathcal{M}_b$
\State $q \gets$ paraphrase$(q_0)$ under the guards of Section~\ref{sec:paraphrase}; $q \gets q_0$ on failure
\State $\tau \gets \mathcal{T}$ solves $(\mathcal{M}_b, q)$ with an edit recipe and no online checker feedback
\If{$s(\mathcal{M}_b,\mathcal{M}',\sigma)=1$} add $(q,\tau)$ to the SFT corpus
\EndIf
\end{algorithmic}
\end{algorithm}

\subsubsection{Synthetic buildings}
\label{sec:buildings}

Training buildings are synthesised from CubiCasa5K floor plans \cite{cubicasa2019}. Each plan becomes a complete IFC model with standard walls (axis and extruded profile), wall joints, openings with doors and windows, spaces whose footprints meet the wall faces, space boundaries, stacked storeys with slabs and property sets; a share of the buildings tiles several plans along a corridor around a vertical core. \claimref{C9004}\evref{E002}\evref{E003}\evref{E004} The model is exported in a randomised \emph{style} that imitates real authoring tools: metres or millimetres; IFC2X3, IFC4 or IFC4X3; rectangular or polyline profiles; Revit- or ArchiCAD-like names; one to four storeys, up to eight in the tiled buildings; wall and space bodies as extrusions, faceted Breps, clipped extrusions or face sets in the shares found in a survey of real exports, which included one of the test projects (Appendix~\ref{sec:supp_inventory}); and, in a fifth of the buildings, the sparse relation set of a Revit export without wall connections and space boundaries.

A survey of the wall openings in the real projects that are not used for evaluation showed that real exports almost never use the plan-profile extrusion of the first generator version: openings are elevation profiles extruded across the wall, rotated or polyline profiles, Brep or face-set bodies, and doors and windows carry shared type geometry through mapped items. A second generator mode draws these opening and fill styles per building with the surveyed shares, so that the training distribution covers the geometry the student meets on real files.

\subsubsection{Task generation}
\label{sec:generation}

Two generators derive tasks and their specifications $\sigma$ from the building model. The \emph{cascade} generator covers four dependency tiers (Table~\ref{tab:taxonomy}). Higher tiers require more dependents to follow the target; preservation protects everything outside the allowed set. Dependents are read from the model's explicit relationships (\texttt{IfcRelVoidsElement}, \texttt{IfcRelFillsElement}, \texttt{IfcRelSpaceBoundary}, \texttt{IfcRelConnectsPathElements}). When an export lacks them, the generator does not infer them from geometry: a wall move on such a model is generated at the tier its explicit relationships support, and the grid templates that need a bounded space are skipped on that model. The verdict therefore depends on the export: the same wall move is a tier-two task in a project with space boundaries and a tier-one task in a project without them, where a bounded space is protected by the preservation gate instead of being required to follow. At tiers one and two, the dependents that the tier does not list may be left inconsistent without failing the task. Quantity sets are not compared by the checkers and are not recomputed by the helpers. A wall move at the deepest tier is also generated at tiers one and two, where space or joint changes are allowed but not required.

The \emph{grid} generator covers creation, update, deletion and retrieval for walls, doors, windows, columns, slabs and spaces. Instructions identify targets \emph{directly} by identifier, \emph{topologically} through relationships, \emph{spatially} through relative geometry, or along a \emph{path} through the building. It also produces basic, geometric, topological and numeric retrieval, quantity take-off, compound instructions and operationally defined conceptual questions. The input building is a derived copy with a few columns added and spaces removed to provide locations for creation tasks. Each task is described by its dependency tier, reference style, operation and target elements. The tier depends on the operation and the target's surroundings: retrieval and property edits have no dependents, most creation, deletion and update tasks occupy the next tier, and wall moves affecting spaces or connected walls occupy the deeper tiers.

\subsubsection{Prompt paraphrase and gates}
\label{sec:paraphrase}

Generated tasks first pass static lint, a scripted oracle and a reference verifier. Lint checks identifiers and parameters. The oracle requires an independently scripted solution to pass all four checkers and a deliberately wrong solution to fail. The reference verifier checks that the instruction identifies its targets uniquely. Tasks with failing oracles are excluded as generator or checker defects.

Kimi K3 \cite{kimik3} paraphrases templated prompts in registers ranging from terse engineering requests to detailed client instructions. The generator retains the specification; the model changes the wording. Guards require every identifier, number and unit to remain verbatim, prohibit new identifiers, preserve retrieval answer formats and bound the rewrite length. Opposed directional terms, including left/right, north/south, clockwise/counter-clockwise, above/below and farthest/nearest, must retain their meaning. A failed rewrite is retried once with the violation specified, then replaced by the original template if it fails again. A separate model call checks semantic fidelity, including whether identifiers still refer to the same elements.

\subsubsection{Corpus and splits}
\label{sec:corpus}

\claimref{C9004}\evref{E002}\evref{E003}\evref{E004} The supervised corpus contains 19,141 strictly successful teacher episodes spanning 3,059 synthetic buildings under Eq.~\eqref{eq:strict}. Buildings are counted by distinct identifiers in the final SFT training split. The separate synthetic validation split contains 2,270 episodes from 340 other buildings. Wall updates form the largest element--operation group in Figure~\ref{fig:overview}. Real buildings are reserved for development and testing. \claimref{C9012}\evref{E019} The test set, IfcEditBench, contains 1,990 tasks from 13 held-out real projects in ifc-bench \cite{ifcbench2026}, excluded from training, pool selection and checkpoint selection. Tasks are drawn according to the training corpus's element--operation distribution, with door, window and opening kinds over-sampled, at least 30 tasks per kind where available, and buildings balanced within each kind; 1,500 tasks cover all families and a supplement of 490 tasks covers doors, windows and openings (the \emph{fill} tasks of the appendix). All pass the gates in Section~\ref{sec:paraphrase}.

\subsection{Training}
\label{sec:training}

IfcEdit has two training stages: imitation of verified teacher episodes (Section~\ref{sec:sft}), followed by GRPO on the resulting student's own rollouts (Section~\ref{sec:grpo}). The second stage draws tasks from a synthetic pool, executes the rollouts in IfcGym and uses strict success as the reward. Distillation \cite{hinton2015kd,studentdistill2025} exposes the student to teacher-visited states; on-policy training evaluates the trajectories produced by the student under the same success criterion.

\subsubsection{Teacher distillation}
\label{sec:sft}

\claimref{C9004}\evref{E002}\evref{E003}\evref{E004} Kimi K3 \cite{kimik3}, used in thinking mode, receives the natural-language task instruction, access to the initial IFC model through the student's tools, and an additional edit recipe in its system prompt. The recipe provides reusable procedures for editing elements, propagating dependent changes and resolving references. For example, it explains how a wall move should update touching wall ends, adjacent space footprints and space boundaries. The teacher resolves the task's targets and parameters from the instruction and the model; the scripted oracle solution is not supplied. Collection allows 40 tool-call rounds, twice the student's budget, with no online checker feedback. After each episode, the four checkers evaluate its final outcome, and only strictly successful episodes are retained. Exception-raising calls are removed from the trace. The student's prompt retains the task instruction and tool documentation but omits the additional recipe. Let $y$ be the resulting message sequence of reasoning, tool calls and tool returns. The set $\mathcal{G}(y)$ contains the positions generated by the model (reasoning and calls) and excludes the environment's tool returns. The student minimises the negative log-likelihood of the generated positions,
\begin{equation}\label{eq:sft}
  \mathcal{L}_{\mathrm{SFT}}(\theta)=-\sum_{(q,y)\in\mathcal{D}}\ \sum_{t\in\mathcal{G}(y)}\log\pi_\theta\bigl(y_t\mid y_{<t},q\bigr),
\end{equation}
which supervises tool selection, argument construction and stopping throughout the episode. \claimref{C9004}\evref{E002}\evref{E003}\evref{E004} The student, Qwen3.5-9B \cite{qwen35,qwen3}, is trained with low-rank adapters \cite{lora2022} on all linear layers of the language model (rank 64, scaling 128), learning rate $10^{-4}$ with cosine decay, three epochs, batch size 16 and sequences up to 32,768 tokens. We use the final checkpoint without selection. The same recipe trains the Qwen3.5-27B student of Section~\ref{sec:experiments}.

\subsubsection{GRPO on the strict reward}
\label{sec:grpo}

The second stage applies Group Relative Policy Optimization (GRPO) \cite{deepseekmath2024,retool2025} to the SFT student with the strict reward as the only signal. For a task $x$ the policy samples a group of $G$ episodes $\{\tau_g\}$ in the sandbox under the evaluation protocol, each scored $r(\tau_g)=s$ by the checkers; the advantage is the group-standardised reward,
\begin{equation}\label{eq:adv}
  A_g=\frac{r(\tau_g)-\mu_r}{\sigma_r},\qquad
  \mu_r=\operatorname*{mean}_{g'} r(\tau_{g'}),\quad
  \sigma_r=\operatorname*{std}_{g'} r(\tau_{g'}),
\end{equation}
so no value network is required, and the policy maximises the clipped surrogate of PPO \cite{ppo2017} with a KL penalty towards the SFT policy $\pi_{\mathrm{ref}}$,
\begin{equation}\label{eq:grpo}
\begin{aligned}
  \mathcal{J}(\theta)=\ &\mathbb{E}\Bigl[\min\bigl(w_g A_g,\\
  &\qquad\quad \mathrm{clip}(w_g,1-\epsilon,1+\epsilon)\,A_g\bigr)\Bigr]\\
  &-\beta\,\mathbb{D}_{\mathrm{KL}}\bigl[\pi_\theta\,\|\,\pi_{\mathrm{ref}}\bigr],
\end{aligned}
\end{equation}
with $w_g=\pi_\theta(\tau_g\mid x)/\pi_{\theta_{\mathrm{old}}}(\tau_g\mid x)$ over the model-generated tokens only. Groups whose episodes all pass or all fail provide no reward-based contrast. Task selection and online filtering favour groups with mixed outcomes.

\paragraph{Task pool} \claimref{C9012}\evref{E019} Tasks are drawn only from synthetic buildings and filtered offline with the starting policy. Each is attempted four times at sampling temperature one. Tasks solved on every attempt are discarded; those solved one to three times are retained. Unsolved tasks are capped at a fifth of the retained pool to limit groups with uniformly failing rollouts.

\paragraph{Optimisation} \claimref{C9012}\evref{E019} Each step samples 64 prompts with $G=8$ rollouts per prompt under the evaluation protocol (budget of 20 rounds, no checker feedback) and updates the adapters in mini-batches of 8 prompts at learning rate $10^{-5}$ in the verl framework \cite{verl2024}. Online group filtering \cite{dapo2025} drops every group whose eight rewards are equal and replaces it with new prompts until the step holds 64 mixed groups, and the KL coefficient $\beta=10^{-2}$ keeps the policy close to the SFT start. The adapters are merged into the rollout weights after every step, so the policy that collects episodes is always the current one. Training runs for 100 steps; the final checkpoint is merged into the base weights and served like the SFT student.

\section{Experiments}
\label{sec:experiments}

\begin{table*}[!t]
 \centering
 \footnotesize
 \setlength{\tabcolsep}{2.5pt}
 \caption{IfcEditBench, 1,990 tasks: strict success (\%) overall, by instruction reference (the seven viewpoint tasks count in All only) and by operation. Blocks: prompted models, the two published frameworks run with the DeepSeek-V4-Flash backbone, and the IfcEdit students. Bold: best among the IfcEdit students.}
 \label{tab:main}
 \begin{tabular*}{\textwidth}{@{\extracolsep{\fill}}l C{24pt} *{4}{C{24pt}} *{5}{C{24pt}}@{}}
 \toprule
 & & \multicolumn{4}{c}{Instruction reference} & \multicolumn{5}{c}{Operation} \\
 \cmidrule(lr){3-6}\cmidrule(lr){7-11}
 Model and training & \multicolumn{1}{C{24pt}}{\scriptsize All} & \multicolumn{1}{C{24pt}}{\scriptsize Direct} & \multicolumn{1}{C{24pt}}{\scriptsize Spatial} & \multicolumn{1}{C{24pt}}{\scriptsize Topol.} & \multicolumn{1}{C{24pt}}{\scriptsize Path} & \multicolumn{1}{C{24pt}}{\scriptsize Create} & \multicolumn{1}{C{24pt}}{\scriptsize Update} & \multicolumn{1}{C{24pt}}{\scriptsize Delete} & \multicolumn{1}{C{24pt}}{\scriptsize Retr.} & \multicolumn{1}{C{24pt}}{\scriptsize Comp.} \\
  & \multicolumn{1}{C{24pt}}{\scriptsize 1{,}990} & \multicolumn{1}{C{24pt}}{\scriptsize 1{,}503} & \multicolumn{1}{C{24pt}}{\scriptsize 217} & \multicolumn{1}{C{24pt}}{\scriptsize 193} & \multicolumn{1}{C{24pt}}{\scriptsize 70} & \multicolumn{1}{C{24pt}}{\scriptsize 355} & \multicolumn{1}{C{24pt}}{\scriptsize 1{,}019} & \multicolumn{1}{C{24pt}}{\scriptsize 331} & \multicolumn{1}{C{24pt}}{\scriptsize 200} & \multicolumn{1}{C{24pt}}{\scriptsize 85} \\
 \midrule
 \rowcolor{black!7}\multicolumn{11}{@{}l}{\textbf{Prompted models}} \\
 DeepSeek-V4-Flash~\cite{deepseekv4} & 61.11 & 56.9 & 76.0 & 81.9 & 50.0 & 61.1 & 62.4 & 51.7 & 82.0 & 32.9 \\
 Gemma 4 31B~\cite{gemma4} & 67.99 & 66.1 & 74.7 & 83.4 & 44.3 & 60.8 & 64.7 & 88.2 & 75.0 & 42.4 \\
 Kimi K3~\cite{kimik3} & 88.04 & 86.6 & 94.0 & 92.7 & 85.7 & 89.9 & 88.1 & 88.5 & 89.5 & 74.1 \\
 Qwen3.5-9B~\cite{qwen35} & 16.08 & 15.4 & 18.9 & 23.8 & 2.9 & 9.3 & 15.6 & 15.7 & 35.0 & 7.1 \\
 Qwen3.5-27B~\cite{qwen35} & 59.35 & 56.2 & 65.9 & 77.7 & 55.7 & 54.6 & 54.9 & 77.9 & 73.5 & 27.1 \\
 \addlinespace[2pt]
 \rowcolor{black!7}\multicolumn{11}{@{}l}{\textbf{Published frameworks}} \\
 MCP4IFC~\cite{mcp4ifc2025} & 39.40 & 39.7 & 39.2 & 49.7 & 10.0 & 12.4 & 36.2 & 64.4 & 69.5 & 22.4 \\
 IFC-Agent~\cite{ifcagent2026} & 13.62 & 11.6 & 10.6 & 35.2 & 8.6 & 0.0 & 14.2 & 0.0 & 63.0 & 0.0 \\
 \addlinespace[2pt]
 \rowcolor{black!7}\multicolumn{11}{@{}l}{\textbf{IfcEdit (ours)}} \\
 Qwen3.5-9B, SFT & 79.15 & 78.7 & 83.4 & 81.9 & 70.0 & 73.0 & 79.7 & 90.3 & 82.0 & 48.2 \\
 Qwen3.5-9B, SFT + GRPO & 86.78 & 86.9 & 86.2 & 90.7 & 77.1 & 87.3 & 86.7 & 94.0 & 86.0 & 60.0 \\
 Qwen3.5-27B, SFT & 88.29 & 87.6 & \tb{89.9} & \tb{92.7} & 85.7 & 85.4 & 89.4 & 94.3 & 88.0 & 64.7 \\
 Qwen3.5-27B, SFT + GRPO & \tb{91.76} & \tb{92.2} & 87.6 & \tb{92.7} & \tb{91.4} & \tb{91.8} & \tb{92.2} & \tb{97.3} & \tb{91.0} & \tb{65.9} \\
 \bottomrule
 \end{tabular*}
\end{table*}

\subsection{Experimental settings}
\label{sec:setup}

\paragraph{Test sets} \claimref{C9012}\evref{E019} IfcEditBench, the test set of Section~\ref{sec:corpus}, holds 1,990 tasks grouped by the taxonomy in Table~\ref{tab:taxonomy}. BIM-Edit \cite{bimedit2026} provides 324 instructions, scored with our calibrated re-implementation of its evaluator. Appendix~\ref{sec:scorer} describes its calibration and protocol reproduction. BIBIMBAP \cite{bibimbap2026} provides 100 instructions with the authors' executable per-task tests.

\begin{figure}[!t]
 \centering
 \includegraphics[width=\linewidth]{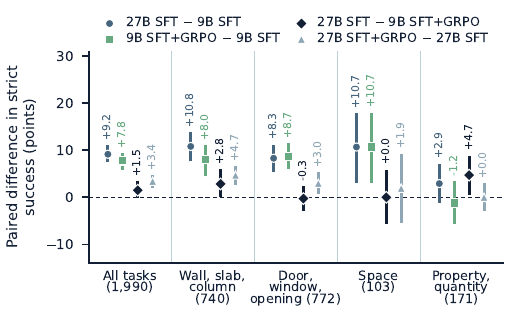}
 \caption{Paired strict-success differences and 95\% cluster-bootstrap intervals for student size, GRPO on each student and the remaining size gap, overall and by target element. Identifier retrieval and compound tasks count in the overall group only; pair counts exclude tasks without an outcome. }
 \label{fig:pairs}
\end{figure}

\paragraph{Models} \claimref{C9012}\evref{E019} \evref{E001} Kimi K3 in thinking mode supplies the teacher episodes (Section~\ref{sec:sft}). The SFT student is Qwen3.5-9B trained with LoRA on the complete supervised corpus, including the second generator's opening and fill styles. The final student adds 100 GRPO steps on a synthetic task pool that includes these styles. The 27B student uses the same SFT data and recipe on Qwen3.5-27B to assess model size; untrained Qwen3.5-27B with the same tools provides its starting point.


\paragraph{Baselines} \claimref{C9012}\evref{E019}\evref{E021}\evref{E020} Prompted comparisons use the students' CodeAct protocol, tools and turn budget without checker feedback: DeepSeek-V4-Flash \cite{deepseekv4} on all test sets, Gemma 4 31B on IfcEditBench, and Kimi K3, the teacher itself, on IfcEditBench and BIBIMBAP. Two published frameworks are evaluated on IfcEditBench with DeepSeek-V4-Flash: the released MCP4IFC tool server \cite{mcp4ifc2025} and our re-implementation of IFC-Agent \cite{ifcagent2026}. We also reproduce BIM-Edit's prompted protocol with Gemma 4 31B to assess our harness and scorer, as detailed in Appendix~\ref{sec:scorer}.

\paragraph{Protocol} \claimref{C9012}\evref{E019} Models are served in bfloat16 with vLLM \cite{vllm2023}, or with SGLang \cite{sglang2024} for Gemma 4 31B and Kimi K3, using common flags, temperature 0, thinking enabled and at most 16,384 output tokens per turn; Kimi K3 runs at its highest reasoning effort. The budget is 20 tool-call rounds, or 25 for BIBIMBAP as prescribed by that benchmark. Unless stated otherwise, models receive no checker feedback. Every model uses a 65,536-token context window, except Gemma 4 31B at 131,072 tokens; an episode that exceeds it counts as a failure. We report strict success in percent on IfcEditBench and the respective benchmark scores for external evaluation. Paired differences carry 95\% cluster-bootstrap intervals \cite{efron1993bootstrap}, using target-element clusters stratified by building and 10,000 resamples. A pair includes only tasks for which both models returned an outcome; an episode without one counts as a failure in the table and is excluded from the pair.

\paragraph{Implementation} \claimref{C9004}\evref{E002}\evref{E003}\evref{E004} Training uses verl \cite{verl2024}, with its supervised trainer for SFT and hybrid engine for GRPO. The latter interleaves generation, sandbox execution and policy updates. Data construction, training and evaluation ran on a cluster of four nodes, each with eight H100 80 GB GPUs.


\begin{figure*}[!t]
 \centering
 \includegraphics[width=\textwidth]{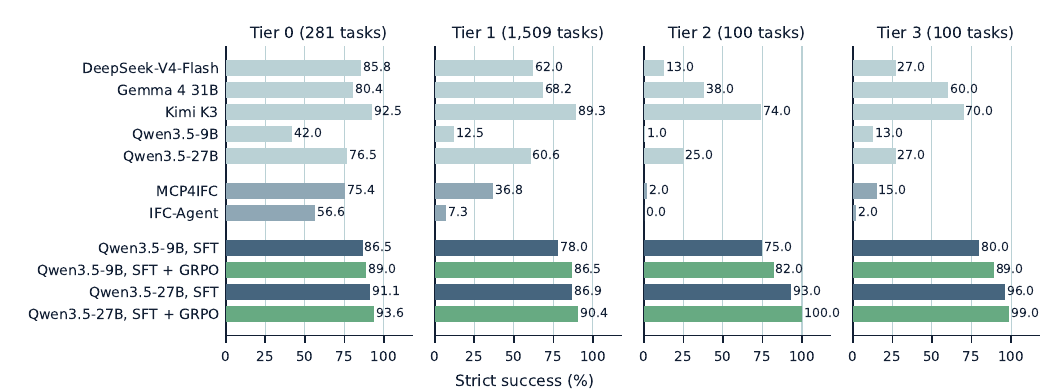}
 \caption{Strict success (\%) by dependency tier for prompted models, published frameworks and trained students. Tiers 2 and 3 hold the wall translations whose bounded spaces or joined walls must follow.}
 \label{fig:tiers}
\end{figure*}

\subsection{Main result}
\label{sec:main}

\begin{table}[t]
 \centering
 \footnotesize
 \caption{IfcEditBench: share of the 1,990 tasks (\%) that pass each checker individually, for prompted models, published frameworks and trained students. Strict success requires all four. Bold: best among the IfcEdit students.}
 \label{tab:checkers}
 \setlength{\tabcolsep}{3pt}
 \begin{tabular*}{\linewidth}{@{\extracolsep{\fill}}l *{4}{C{27pt}}@{}}
 \toprule
 Model and training & \multicolumn{1}{C{27pt}}{\scriptsize Valid.} & \multicolumn{1}{C{27pt}}{\scriptsize Intent} & \multicolumn{1}{C{27pt}}{\scriptsize Dep.} & \multicolumn{1}{C{27pt}}{\scriptsize Pres.} \\
 \midrule
 \rowcolor{black!7}\multicolumn{5}{@{}l}{\textbf{Prompted models}} \\
 DeepSeek-V4-Flash~\cite{deepseekv4} & 96.0 & 68.3 & 80.4 & 96.4 \\
 Gemma 4 31B~\cite{gemma4} & 96.5 & 77.5 & 90.3 & 93.0 \\
 Kimi K3~\cite{kimik3} & 99.4 & 91.8 & 95.7 & 96.4 \\
 Qwen3.5-9B~\cite{qwen35} & 97.8 & 19.5 & 56.7 & 97.7 \\
 Qwen3.5-27B~\cite{qwen35} & 98.8 & 65.6 & 80.9 & 95.7 \\
 \addlinespace[2pt]
 \rowcolor{black!7}\multicolumn{5}{@{}l}{\textbf{Published frameworks}} \\
 MCP4IFC~\cite{mcp4ifc2025} & 85.1 & 50.3 & 76.0 & 88.5 \\
 IFC-Agent~\cite{ifcagent2026} & 90.3 & 24.4 & 56.1 & 75.8 \\
 \addlinespace[2pt]
 \rowcolor{black!7}\multicolumn{5}{@{}l}{\textbf{IfcEdit (ours)}} \\
 Qwen3.5-9B, SFT & \tb{99.2} & 80.9 & 92.6 & 96.8 \\
 Qwen3.5-9B, SFT + GRPO & 99.1 & 87.4 & 95.6 & \tb{98.0} \\
 Qwen3.5-27B, SFT & 98.7 & 89.6 & 96.7 & 97.3 \\
 Qwen3.5-27B, SFT + GRPO & 98.8 & \tb{92.5} & \tb{98.0} & 97.5 \\
 \bottomrule
 \end{tabular*}
\end{table}

\begin{table*}[t]
 \centering
 \footnotesize
 \setlength{\tabcolsep}{4pt}
 \caption{BIBIMBAP, 100 tasks: score (\%) by category, operation and overall. Rows marked $^\dagger$ are reported by the benchmark \cite{bibimbap2026}. Blocks distinguish prompted models and IfcEdit students; bold marks the best value among the rows we evaluated.}
 \label{tab:bibimbap}
 \begin{tabular*}{\textwidth}{@{\extracolsep{\fill}}l *{5}{C{31pt}} *{4}{C{31pt}} C{31pt}@{}}
 \toprule
 & \multicolumn{5}{c}{Category (20 tasks each)} & \multicolumn{4}{c}{Operation} & \\
 \cmidrule(lr){2-6}\cmidrule(lr){7-10}
 Model & \multicolumn{1}{C{31pt}}{\scriptsize Direct} & \multicolumn{1}{C{31pt}}{\scriptsize Spatial} & \multicolumn{1}{C{31pt}}{\scriptsize Topol.} & \multicolumn{1}{C{31pt}}{\scriptsize Numeric} & \multicolumn{1}{C{31pt}}{\scriptsize Concept.} & \multicolumn{1}{C{31pt}}{\scriptsize Create} & \multicolumn{1}{C{31pt}}{\scriptsize Read} & \multicolumn{1}{C{31pt}}{\scriptsize Update} & \multicolumn{1}{C{31pt}}{\scriptsize Delete} & \multicolumn{1}{C{31pt}}{\scriptsize All} \\
 \midrule
 \rowcolor{black!7}\multicolumn{11}{@{}l}{\textbf{Prompted models}} \\
 Claude Opus 4.6~\cite{claudeopus46}$^\dagger$ & 65.83 & 63.89 & 53.06 & 38.33 & 30.00 & 81.1 & 47.3 & 61.5 & 18.9 & 50.22 \\
 GPT-5.4~\cite{gpt54}$^\dagger$ & 62.15 & 42.08 & 20.56 & 26.67 & 5.00 & 56.0 & 22.4 & 50.4 & 20.0 & 31.29 \\
 Gemini 3 Flash~\cite{gemini3flash}$^\dagger$ & 65.42 & 40.97 & 21.11 & 20.00 & 6.67 & 45.6 & 20.6 & 46.7 & 37.8 & 30.83 \\
 GPT-5.2~\cite{gpt52}$^\dagger$ & 52.50 & 34.58 & 35.42 & 5.00 & 25.00 & 58.9 & 18.2 & 51.1 & 26.7 & 30.50 \\
 GPT-5 Mini~\cite{gpt5}$^\dagger$ & 50.94 & 25.83 & 25.83 & 0.00 & 25.00 & 31.2 & 10.9 & 55.5 & 43.3 & 25.00 \\
 GPT-4.1~\cite{gpt41}$^\dagger$ & 38.33 & 30.00 & 31.25 & 0.00 & 20.00 & 29.4 & 9.0 & 60.0 & 36.6 & 23.92 \\
 DeepSeek-V4-Flash~\cite{deepseekv4} & 68.33 & 55.83 & 53.33 & 35.00 & 25.00 & 76.7 & 41.8 & 73.3 & 13.3 & 47.50 \\
 Kimi K3~\cite{kimik3} & 77.50 & 67.08 & 75.00 & 35.00 & \tb{60.00} & 91.7 & \tb{60.0} & 71.1 & \tb{36.7} & 62.92 \\
 Qwen3.5-9B~\cite{qwen35} & 45.00 & 40.00 & 26.67 & 25.00 & 15.00 & 16.7 & 29.1 & 62.2 & 16.7 & 30.33 \\
 Qwen3.5-27B~\cite{qwen35} & 66.25 & 60.42 & 66.67 & 35.00 & 35.00 & 73.3 & 49.1 & 71.1 & 26.7 & 52.67 \\
 \addlinespace[2pt]
 \rowcolor{black!7}\multicolumn{11}{@{}l}{\textbf{IfcEdit (ours)}} \\
 Qwen3.5-9B, SFT & 77.50 & 57.08 & 65.83 & 30.00 & 30.00 & 91.7 & 41.8 & 68.9 & 33.3 & 52.08 \\
 Qwen3.5-9B, SFT + GRPO & \tb{79.17} & 60.00 & 73.33 & 30.00 & 30.00 & \tb{100.0} & 41.8 & 73.3 & \tb{36.7} & 54.50 \\
 Qwen3.5-27B, SFT & 76.67 & 66.67 & \tb{75.83} & 30.00 & 50.00 & \tb{100.0} & 52.7 & \tb{75.6} & 30.0 & 59.83 \\
 Qwen3.5-27B, SFT + GRPO & \tb{79.17} & \tb{71.67} & 73.33 & \tb{45.00} & 50.00 & \tb{100.0} & \tb{60.0} & \tb{75.6} & 30.0 & \tb{63.83} \\
 \bottomrule
 \end{tabular*}
\end{table*}

GRPO improves the 9B student and narrows its gap to the 27B SFT model. \claimref{C9012}\evref{E019} On the 1,990 IfcEditBench tasks, the final student reaches 86.78\% compared with 79.15\% for its SFT start, 88.29\% for the 27B SFT student, and 59.35\% and 16.08\% for the untrained models (Table~\ref{tab:main}). \claimref{C9012}\evref{E019}\evref{E021} By instruction reference, the final student resolves 86.9\% of direct, 86.2\% of spatial and 90.7\% of topological references, and 77.1\% of the 70 path references, the hardest group for every model but the 27B GRPO student. By operation, GRPO lifts the SFT student most on creations, from 73.0\% to 87.3\%, and on compound edits, from 48.2\% to 60.0\%, while deletions already reach 90.3\% after SFT. The prompted models show a different profile: DeepSeek-V4-Flash resolves 81.9\% of topological references but completes 51.7\% of deletions and 32.9\% of compound edits, and the two published frameworks stay below 13\% on creations and on path references. \claimref{C9013}\evref{E024} The 27B GRPO student reaches 91.76\% and leads the students in every column except spatial and topological references. The 27B SFT student leads on spatial references; the two tie on topological references. The teacher exceeds the 27B GRPO student on spatial references and compound edits.

The paired comparisons in Figure~\ref{fig:pairs} distinguish model size from the reinforcement stage. Applying the same SFT recipe and data to the larger model gives $+9.15$ points, with interval $[7.42, 11.05]$; the difference is positive in 12 of the 13 projects and zero in the remaining project. GRPO adds $+7.75$ points $[5.79, 9.47]$. The larger SFT model retains a lead of $+1.51$ points $[-0.05, 3.48]$ over the final student, while the door, window and opening difference is $-0.26$ $[-2.99, 2.41]$; both intervals include zero. Improvements are prominent for spaces and for walls, slabs and columns; the size and reinforcement contrasts on property and quantity tasks are not clearly separated from zero. \claimref{C9013}\evref{E024} GRPO on the 27B SFT student, with the same recipe, reaches 91.76\% (Table~\ref{tab:main}); it adds $+3.42$ $[1.99, 4.71]$ over its SFT start.

\claimref{C9014}\evref{E019}\evref{E021}\evref{E022}\evref{E024} Figure~\ref{fig:tiers} splits the results in Table~\ref{tab:main} by the dependency tiers defined in Table~\ref{tab:taxonomy}. Apart from the untrained smaller model and the teacher, the prompted models solve 76.5\% to 85.8\% of the tier-0 tasks, which touch one element, but 13.0\% to 38.0\% of the tier-2 tasks, where a translated wall must carry its space boundaries, and the two frameworks fall to 2.0\% and 0.0\% there. The teacher solves 92.5\% of the tier-0 and 74.0\% of the tier-2 tasks. Every trained student stays at or above 75\% on all four tiers; the 27B GRPO student reaches 100.0\% and 99.0\% on tiers 2 and 3. The largest differences between trained and prompted models occur on these dependent edits. The values of both figures are tabulated in Appendix~\ref{sec:supp_tiers}.

The checker breakdown (Table~\ref{tab:checkers}) shows higher intent, dependency and preservation pass rates after GRPO. Validity is already high and changes little, while intent pass rates differ most across models. The larger student after GRPO has the highest intent pass rate of all students.

\paragraph{Prompted and framework references} \claimref{C9012}\evref{E019}\evref{E021}\evref{E020} The trained students outperform every prompted comparison other than the teacher under the shared tool interface. DeepSeek-V4-Flash reaches 61.11\% strict success, about two points above the untrained larger model and $25.68$ points $[23.12, 28.10]$ below the final student. Gemma 4 31B reaches 67.99\%, leaving a gap of $18.79$ points $[16.35, 20.90]$. The teacher, Kimi K3, reaches 88.04\% under the student's budget of 20 rounds, $1.16$ points $[-0.73, 3.36]$ above the final student and $3.72$ points $[1.78, 5.59]$ below the 27B GRPO student; the interval includes zero for the 9B comparison and excludes zero for the 27B comparison. Twenty-eight of the teacher's episodes exceeded the context window and count as failures. The published frameworks score lower with the same DeepSeek-V4-Flash backbone: 39.40\% for MCP4IFC and 13.62\% for IFC-Agent, which averages 130 tool calls per task. Neither provides a dependency-aware edit primitive. MCP4IFC's reported result is the better of two tool-output caps; context overflow at the 65,536-token limit affects 5\% of tasks with the tighter cap and 27\% with the looser cap. The frameworks differ in both tool interface and execution constraints; the prompted-model comparisons hold the interface fixed to assess training.

\subsection{External benchmarks}
\label{sec:external}

\begin{table}[!t]
 \centering
 \footnotesize
 \caption{BIM-Edit, 324 tasks: average final, geometry, semantics and topology scores. Rows marked $^\dagger$ are reported by the benchmark \cite{bimedit2026} with the official scorer; the other rows use our calibrated re-implementation. Blocks distinguish prompted models and IfcEdit students; bold marks the best value among the rows we evaluated.}
 \label{tab:bimedit}
 \resizebox{\linewidth}{!}{%
 \begin{tabular}{l *{4}{C{30pt}}}
 \toprule
 Model & \multicolumn{1}{C{30pt}}{Final} & \multicolumn{1}{C{30pt}}{Geom.} & \multicolumn{1}{C{30pt}}{Sem.} & \multicolumn{1}{C{30pt}}{Topo.} \\
 \midrule
 \rowcolor{black!7}\multicolumn{5}{@{}l}{\textbf{Prompted models}} \\
 Gemini 3 Flash~\cite{gemini3flash}$^\dagger$ & 49.48 & 68.87 & 41.81 & 37.77 \\
 Qwen 3.6 Plus~\cite{qwen36plus}$^\dagger$ & 47.63 & 59.21 & 35.93 & 47.77 \\
 Claude Sonnet 4.6~\cite{claudesonnet46}$^\dagger$ & 45.31 & 55.90 & 33.85 & 46.19 \\
 GPT-5.4 Pro~\cite{gpt54}$^\dagger$ & 43.94 & 50.14 & 38.69 & 42.97 \\
 DeepSeek V3.2~\cite{deepseekv32}$^\dagger$ & 43.21 & 50.97 & 34.80 & 43.86 \\
 GPT-5.4 Mini~\cite{gpt54}$^\dagger$ & 39.79 & 45.18 & 38.80 & 35.38 \\
 Gemma 4 31B~\cite{gemma4}$^\dagger$ & 37.54 & 44.94 & 29.94 & 37.75 \\
 DeepSeek-V4-Flash~\cite{deepseekv4} & 41.63 & 46.74 & 32.56 & 45.60 \\
 Qwen3.5-9B~\cite{qwen35} & 21.53 & 10.72 & 29.40 & 24.46 \\
 Qwen3.5-27B~\cite{qwen35} & 35.34 & 37.21 & 31.68 & 37.14 \\
 \addlinespace[2pt]
 \rowcolor{black!7}\multicolumn{5}{@{}l}{\textbf{IfcEdit (ours)}} \\
 Qwen3.5-9B, SFT & 50.46 & 61.03 & \tb{36.06} & 54.29 \\
 Qwen3.5-9B, SFT + GRPO & 55.19 & 71.58 & 34.88 & 59.12 \\
 Qwen3.5-27B, SFT & 54.76 & 69.95 & 33.66 & 60.68 \\
 Qwen3.5-27B, SFT + GRPO & \tb{56.94} & \tb{75.50} & 32.82 & \tb{62.50} \\
 \bottomrule
 \end{tabular}}
\end{table}

\begin{figure}[!t]
 \centering
 \includegraphics[width=0.9\linewidth]{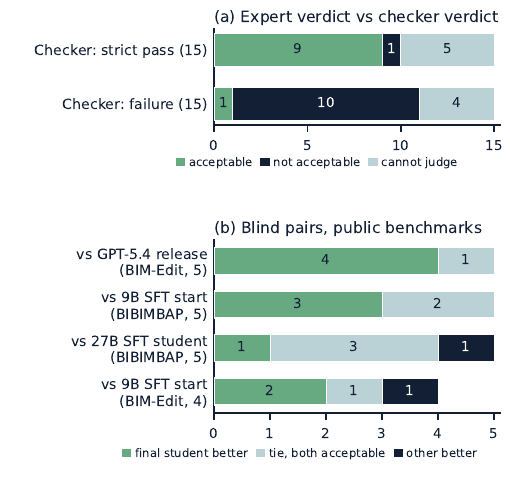}
 \caption{Expert evaluation by one practitioner. (a) Verdict on 30 IfcEditBench edits of the final student, 15 strict passes and 15 failures under the four checkers, judged from before-and-after views. (b) Blind pairwise preference between the final student and another output on the public benchmarks.}
 \label{fig:humaneval}
\end{figure}

\begin{figure*}[!t]
 \centering
 \includegraphics[width=\textwidth]{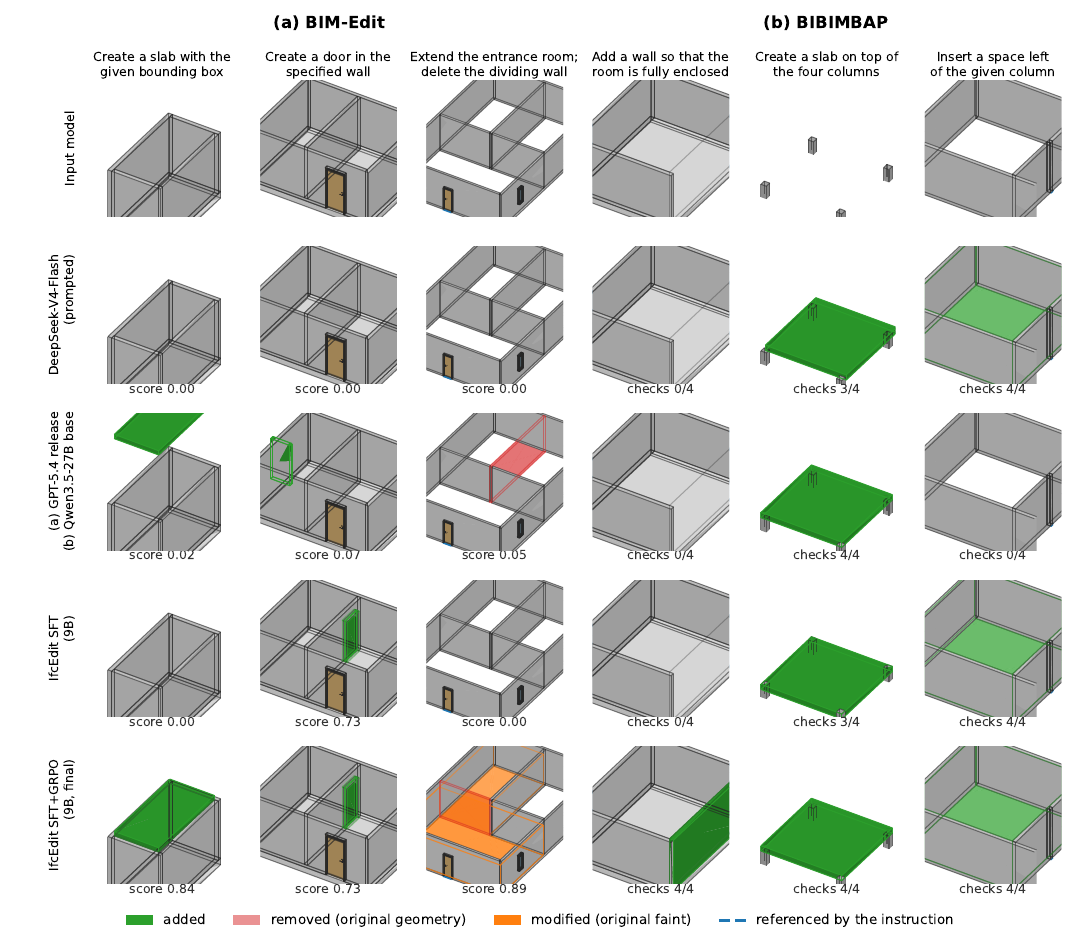}
 \caption{Six public-benchmark edits (columns) across comparison models and trained students (rows). Section views mark additions in green, removals in red using original geometry, and modifications in orange. Labels give BIM-Edit scores under the re-implemented evaluator or the number of BIBIMBAP checks passed.}
 \label{fig:examples}
\end{figure*}

\claimref{C9010}\evref{E019}\evref{E021} The trained students improve over their starting models on both external benchmarks. On BIBIMBAP (Table~\ref{tab:bibimbap}), the final student scores 54.50, its SFT start 52.08, the larger SFT student 59.83 and the untrained larger model 52.67; the final and larger students exceed the published Claude Opus 4.6 score of 50.2\% \cite{bibimbap2026}. The teacher, Kimi K3, scores 62.92 there, $8.42$ points $[1.42, 15.75]$ above the final student and $0.92$ points $[-8.00, 6.00]$ below the 27B GRPO student, with an interval that includes zero. On BIM-Edit (Table~\ref{tab:bimedit}), the final student scores 55.19, its SFT start 50.46, the larger SFT student 54.76 and the untrained larger model 35.34. GRPO adds $+4.73$ points $[2.77, 6.70]$, similar to the size effect of $+4.30$ $[2.11, 6.57]$. DeepSeek-V4-Flash scores 47.50 on BIBIMBAP and 41.63 on BIM-Edit. Our BIM-Edit scorer differs from the official evaluator used for published rows, so those cross-block comparisons remain approximate; Appendix~\ref{sec:scorer} reports the reproduction.

\claimref{C9013}\evref{E024} The 27B GRPO student scores 63.83 on BIBIMBAP and 56.94 on BIM-Edit, against 59.83 and 54.76 for its SFT start.

\paragraph{Expert evaluation} \claimref{C9015}\evref{E019} A practitioner reviewed 30 IfcEditBench edits of the final student: 15 strict passes and 15 failures under the checkers. The review used before-and-after views with the changed elements highlighted. The same practitioner also reviewed 20 blind pairs on the public benchmarks (Figure~\ref{fig:humaneval}); one pair was left unanswered because the benchmark instruction contains a malformed coordinate. The expert judged 21 of the edits and marked 9 as not judgeable from the views; on the 21, the verdict agrees with the four-gate verdict in 19 cases, rejecting 10 of the 11 checker failures and accepting 9 of the 10 passes. In the blind pairs the final student was preferred in 10 and tied as acceptable in 7, against 2 for the other output; against the released GPT-5.4 Pro outputs on BIM-Edit it was preferred in 4 of 5 pairs. Where a side was preferred, the choice agrees with the sign of the benchmark score in 10 of 12 pairs.

\begin{samepage}
Figure~\ref{fig:examples} compares six creation, deletion and room-level edits from the final student, its SFT start and the other evaluated systems on the two benchmarks. Further IfcEditBench edits and benchmark pairs are shown in Appendix~\ref{sec:supp_examples}.\par
\end{samepage}

\subsection{Ablations}
\label{sec:ablations}

\begin{figure}[!t]
 \centering
 \includegraphics[width=\linewidth]{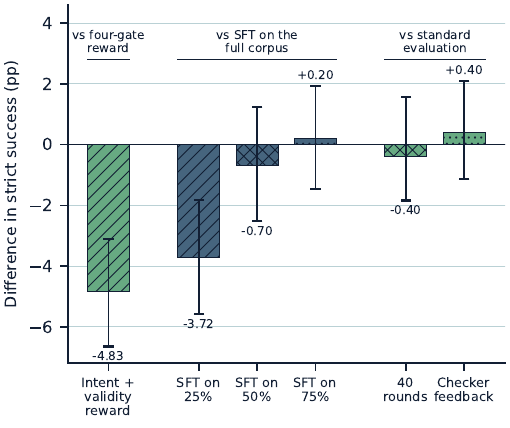}
 \caption{Ablations of the 9B student as paired differences in strict success on IfcEditBench with 95\% cluster-bootstrap intervals: the intent-and-validity reward against the four-gate reward, both after GRPO; SFT on nested fractions of the corpus against SFT on the full corpus; and the final student with a doubled turn budget or with checker feedback against its standard evaluation.}
 \label{fig:ablations}
\end{figure}

\claimref{C9012}\evref{E019}\evref{E022}\evref{E023} Figure~\ref{fig:ablations} compares reward terms, supervised data scale and evaluation-time settings.

\paragraph{Reward terms} \claimref{C9012}\evref{E019}\evref{E022} The ablation repeats GRPO with reward 1 when intent and validity pass and 0 otherwise, logging the other gates without rewarding them. The intent-and-validity policy reaches 82.01\% on IfcEditBench, 52.14 on BIM-Edit and 57.75 on BIBIMBAP; its paired differences from the four-gate policy are $-4.83$ $[-6.65, -3.10]$, $-3.06$ $[-4.89, -1.29]$ and $+3.25$ $[-0.25, 7.42]$. Adding dependency and preservation to the reward improves IfcEditBench performance.

\paragraph{Data scale} \claimref{C9012}\evref{E019}\evref{E023} Much of the IfcEditBench performance is obtained from a fraction of the supervised corpus. The ablation uses nested random quarters with the same trainer seed. The 25\%, 50\% and 75\% students reach 75.38\%, 78.39\% and 79.35\% on IfcEditBench, paired differences from the full-corpus student of $-3.72$ $[-5.58, -1.82]$, $-0.70$ $[-2.52, 1.23]$, $+0.20$ $[-1.46, 1.92]$. On BIM-Edit they score 47.41, 49.24 and 52.94 against 50.46, differences of $-3.05$ $[-5.50, -0.63]$, $-1.22$ $[-3.60, 1.15]$, $+2.48$ $[-0.07, 5.05]$; on BIBIMBAP 47.00, 51.67 and 51.58 against 52.08, differences of $-5.08$ $[-11.08, 0.58]$, $-0.42$ $[-6.33, 5.58]$, $-0.50$ $[-6.42, 5.33]$.

\paragraph{Budget and feedback} \claimref{C9012}\evref{E019}\evref{E022} Doubling the budget to 40 rounds leaves the final student at 86.38\% on IfcEditBench, a paired difference of $-0.40$ $[-1.84, 1.56]$ from its standard evaluation; returning the four checker verdicts after each executed step gives 87.29\%, $+0.40$ $[-1.14, 2.09]$.

\begin{figure}[!htb]
 \centering
 \includegraphics[width=\linewidth]{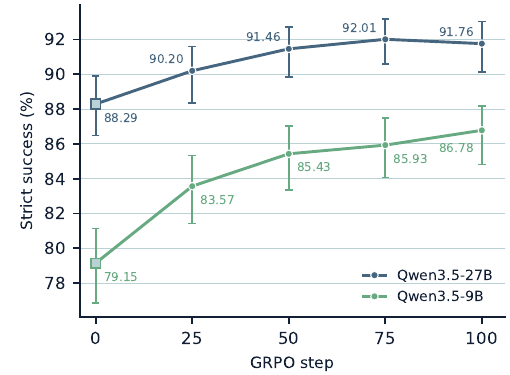}
 \caption{GRPO on the 27B and 9B SFT students: strict success (\%) on the 1,990 IfcEditBench tasks at the exported checkpoints. Step 0 is the SFT student (square); step 100 the final GRPO student. Bars give 95\% cluster-bootstrap intervals. }
 \label{fig:curve}
\end{figure}

\paragraph{Training steps} \claimref{C9013}\evref{E024}\evref{E022} Figure~\ref{fig:curve} follows both GRPO runs through their exported checkpoints. For the 27B student, IfcEditBench strict success rises from 88.29\% to 90.20\% at step 25 and 91.46\% at step 50, then stays between 91.76\% and 92.01\%. For the 9B student it rises from 79.15\% to 83.57\% at step 25 and 85.43\% at step 50, reaches 85.93\% at step 75 and 86.78\% at step 100. Most of the GRPO gain occurs in the first half of training at both scales. Appendix~\ref{sec:supp_curve} lists every exported 27B checkpoint with its benchmark scores.

\section{Discussion}
\label{sec:discussion}

The trained students complete dependent edits, such as moving a wall with its space boundaries and joined walls, more often than the other prompted models. The small student approaches its teacher's performance with the same tools and a fraction of its parameters. An in-house deployment could use such a model for routine edits while keeping project data local. For tasks with an executable specification, the four checkers could also control whether an authoring tool accepts an edit: validity, intent, dependency and preservation must all pass. Free-form instructions would first need to be converted into specifications.

Engineers can inspect and extend the domain rules in the task generator and checkers \cite{wang2023learning}, while the policy learns to satisfy them through the tools. Executable specifications allow training tasks to be generated on synthetic buildings and the same checks to evaluate edits on real projects, without target files. Applying this approach to other structured artefacts, such as parametric CAD assemblies, structural analysis models, product lifecycle data or schema-bound databases, would require task generators with checkable specifications and a sandbox for the agent's code.

Strict success measures compliance with the task specification. Validity checks for new schema errors and geometry failures relative to the input; dependent elements are checked against the rules supplied by the generator and the relationships present in the export. Broader conformance and consistency require additional checks. Coverage of real export geometry depends on the synthetic styles used for training. Projects and generation seeds are held out, but task templates are shared between training and test, leaving transfer to unseen templates untested. The comparisons include two frontier models, one of them the teacher, one open model and two published frameworks on one backbone. The agent edits the element and property kinds represented in its training data. Extending the dependency rules to relations missing from the current specifications and evaluating on held-out templates would test a wider range of edits.

\section{Conclusion}
\label{sec:conclusion}

We train IFC editing agents against task specifications that check the requested change, its dependents and the preservation of unrelated entities. IfcGym provides a replayable sandbox and four checkers for validity, intent, dependency and preservation. Their conjunction selects training episodes, supplies the reinforcement reward and measures strict success without comparing with a target file.

With tool access held fixed on IfcEditBench, verified teacher distillation followed by GRPO brings the small open model close to its teacher's performance; the same recipe takes the larger student beyond the teacher. The gain from reinforcement is comparable to the gain from increasing student size. Both students improve over their starting models on the public benchmarks and outperform the other prompted models and evaluated frameworks on IfcEditBench.

\section*{Acknowledgements}
This work was supported by the National Natural Science Foundation of China (Grant No. 52568010).

\section*{Code and data availability}
The IfcGym environment and the IfcEdit code (agent, evaluation runners, training recipe) are released under the Apache License.\footnote{\url{https://github.com/wbw520/ifcgym}} IfcEditBench is available with the real projects under their original licences (Creative Commons Attribution, MIT, GPL and the KIT terms of use, as listed with each project) and with the task specifications under Creative Commons Attribution.\footnote{\url{https://huggingface.co/datasets/wbw520wbw520/ifceditbench}} The synthetic training buildings, the generated tasks, the supervised corpus and the reinforcement task pool are available under Creative Commons Attribution-NonCommercial, as the floor plans derive from CubiCasa5K.\footnote{\url{https://huggingface.co/datasets/wbw520wbw520/ifcgym-synthetic}} BIM-Edit and BIBIMBAP are distributed by their authors.

{
    \small
    \bibliographystyle{ieeenat_fullname}
    \bibliography{references}
}

\clearpage
\onecolumn
\setcounter{section}{0}
\renewcommand{\thesection}{\Alph{section}}
\begin{center}
  {\Large\bfseries Supplementary Material}\\[6pt]
  {\large Learning to Edit IFC Building Models: Verifiable Checkers as the Training Signal for Dependency-Aware Editing Agents}
\end{center}
\vspace{1em}
\section{The test projects}
\label{sec:supp_inventory}

IfcEditBench draws its 1,990 tasks from 13 real projects of the ifc-bench collection \cite{ifcbench2026}, none of which appears in training, pool selection or checkpoint selection; the seven other projects of the collection on disk served as development projects. Tables~\ref{tab:supp_inventory} and \ref{tab:supp_inventory_b} list, for every test project, the schema, length unit and authoring tool recorded in the file header, the element counts, the representation types of the wall bodies, the relationship classes the generators read, and the number of tasks by dependency tier. The projects span three schema versions, three length units, Revit, ArchiCAD and a Synchro re-export, and wall bodies from plain extrusions to clipped extrusions, faceted Breps and polygonal face sets; six projects carry neither wall connections nor space boundaries, two carry wall connections only, and one (a Synchro re-export) carries no openings, fills or spaces at all. The cascade templates read a project's \texttt{arc.ifc} directly (603 tasks); the grid templates read a derived copy, \texttt{arc\_v4.ifc}, that differs from it only by three to six added columns and one or two removed spaces, which provide locations and targets for creation tasks (1,387 tasks).

\begin{table}[!htbp]
 \centering
 \scriptsize
 \setlength{\tabcolsep}{2pt}
 \renewcommand{\arraystretch}{1.1}
 \caption{The 13 test projects (a): schema, length unit and authoring tool from the file header, and element counts in the original \texttt{arc.ifc}.}
 \label{tab:supp_inventory}
 \begin{tabular}{@{}lllp{0.15\textwidth}rrrrrrrrr@{}}
 \toprule
 Project & Schema & Unit & Authoring tool & Products & Walls & Doors & Windows & Openings & Slabs & Columns & Spaces & Storeys \\
 \midrule
 4351 & IFC2X3 & ft & Revit Architecture 2011 & 441 & 43 & 17 & 3 & 32 & 6 & 0 & 6 & 5 \\
 city\_house\_munich & IFC4X3 & m & Revit 2025 & 293 & 83 & 16 & 13 & 33 & 19 & 0 & 19 & 15 \\
 dental\_clinic & IFC2X3 & m & Revit Architecture 2011 & 3,298 & 1,080 & 254 & 58 & 403 & 3 & 0 & 269 & 4 \\
 digital\_hub & IFC4 & m & Revit 2019 & 1,026 & 178 & 64 & 47 & 244 & 23 & 62 & 64 & 3 \\
 duplex & IFC2X3 & m & Revit Architecture 2011 & 295 & 57 & 14 & 24 & 50 & 21 & 0 & 21 & 4 \\
 fantasy\_hotel\_1 & IFC4 & m & Revit 2024 & 1,082 & 53 & 25 & 73 & 138 & 49 & 39 & 26 & 4 \\
 fantasy\_hotel\_2 & IFC4 & mm & Revit 2024 & 1,571 & 118 & 72 & 190 & 285 & 78 & 86 & 20 & 6 \\
 fantasy\_office\_building\_1 & IFC4 & m & Revit 2025 & 1,846 & 64 & 34 & 42 & 120 & 18 & 34 & 44 & 5 \\
 fantasy\_office\_building\_3 & IFC4 & m & Revit 2024 & 1,460 & 223 & 75 & 167 & 304 & 17 & 20 & 87 & 6 \\
 molio & IFC2X3 & mm & ArchiCAD 25 & 4,192 & 189 & 119 & 119 & 245 & 53 & 14 & 117 & 10 \\
 samuel\_macalister & IFC4 & mm & Revit 2014 & 524 & 48 & 16 & 17 & 33 & 12 & 3 & 14 & 6 \\
 schependomlaan & IFC2X3 & mm & Synchro re-export & 3,508 & 880 & 205 & 259 & 0 & 279 & 23 & 0 & 1 \\
 smiley\_west & IFC4 & m & ArchiCAD 20 & 1,272 & 281 & 170 & 80 & 280 & 120 & 20 & 140 & 5 \\
 \midrule
 All 13 &  &  &  & 20,808 & 3,297 & 1,081 & 1,092 & 2,167 & 698 & 301 & 827 & 74 \\
 \bottomrule
 \end{tabular}
\end{table}

\begin{table}[!htbp]
 \centering
 \footnotesize
 \setlength{\tabcolsep}{3.5pt}
 \renewcommand{\arraystretch}{1.1}
 \caption{The 13 test projects (b): share of walls by Body representation type (S swept solid, C clipping result, B faceted Brep, F polygonal face set, M surface model, I mapped item, ``--'' axis only), counts of path connections, space boundaries, voids and fills relationships, and IfcEditBench tasks by dependency tier.}
 \label{tab:supp_inventory_b}
 \begin{tabular}{@{}lp{0.2\textwidth}rrrrcr@{}}
 \toprule
 Project & Wall bodies & Connections & Boundaries & Voids & Fills & Tasks T0/T1/T2/T3 & Total \\
 \midrule
 4351 & S 58, C 37, M 5 & 65 & 142 & 32 & 20 & 22/153/16/16 & 207 \\
 city\_house\_munich & S 53, C 47 & 93 & 0 & 33 & 26 & 22/174/0/37 & 233 \\
 dental\_clinic & S 96, C 2, M 1 & 2,103 & 3,124 & 403 & 302 & 23/184/32/16 & 255 \\
 digital\_hub & S 81, F 19 & 0 & 0 & 244 & 109 & 14/69/0/0 & 83 \\
 duplex & S 91, C 7, M 2 & 82 & 265 & 50 & 38 & 18/166/15/14 & 213 \\
 fantasy\_hotel\_1 & S 64, F 36 & 0 & 0 & 138 & 98 & 15/68/0/0 & 83 \\
 fantasy\_hotel\_2 & S 92, F 8 & 0 & 0 & 285 & 262 & 17/65/0/0 & 82 \\
 fantasy\_office\_building\_1 & S 91, F 9 & 0 & 0 & 120 & 72 & 24/64/0/0 & 88 \\
 fantasy\_office\_building\_3 & S 27, F 73 & 0 & 0 & 304 & 242 & 22/61/0/0 & 83 \\
 molio & B 52, -- 48 & 282 & 1,973 & 245 & 238 & 31/136/6/0 & 173 \\
 samuel\_macalister & S 54, C 35, B 10 & 75 & 0 & 33 & 24 & 19/116/0/0 & 135 \\
 schependomlaan & B 100 & 0 & 0 & 0 & 0 & 34/67/0/0 & 101 \\
 smiley\_west & S 96, I 4 & 369 & 1,689 & 280 & 250 & 20/186/31/17 & 254 \\
 \midrule
 All 13 & S 56, B 30, F 7, C 3 & 3,069 & 7,193 & 2,167 & 1,681 & 281/1,509/100/100 & 1,990 \\
 \bottomrule
 \end{tabular}
\end{table}

\paragraph{Tasks and relationships}
The relationships present in each file determine which dependency tiers the generators can construct (Section~\ref{sec:supp_checkers}, Table~\ref{tab:supp_inventory_b}). Tier-2 tasks, which require bounded spaces to follow a moved wall, exist only in the five projects with space boundaries; tier-3 tasks, which require joined walls to stay joined, in five of the seven projects with path connections, and in city\_house\_munich, which exports connections but no boundaries, all 37 deep-tier tasks are joint tasks without a space rule. The six projects without either relationship class, and samuel\_macalister, which has no deep-tier task, contribute tier-0 and tier-1 tasks only: attribute and property edits, deletions, creations, retrievals and the fill-supplement tasks on doors, windows and openings. By reference style, 815 tasks name the target by identifier, 217 by relative geometry, 193 through relationships, 70 along a route and 7 from a viewpoint; 85 are compound instructions and the 603 cascade tasks use templated direct references.

\paragraph{Sources and licences}
The projects come from the public ifc-bench collection; their individual licences, recorded with each project, are CC BY 4.0 (city\_house\_munich, dental\_clinic, duplex, molio, schependomlaan), MIT (digital\_hub and the four fantasy projects of TUM student exercises), GPL v3 (4351 and samuel\_macalister, from the BIMserver test files) and the KIT terms of unrestricted use with a citation request (smiley\_west). The IfcEditBench release retains each project's original licence.

\paragraph{The export-style surveys}
Two surveys set the export-style shares of the synthetic training buildings. The body-style shares (extrusion, faceted Brep, clipped extrusion, face set, and the share of sparse exports) were read in October 2026 from 33 real models on disk: the seven development projects of ifc-bench, the two BIBIMBAP models and 24 BIM-Edit models, and also schependomlaan, which at that time was the only test project on disk; its 880 Brep walls contributed to the Brep share. No task, instruction or checker output of that project entered training; the share of Brep wall bodies in the synthetic buildings reflects, in part, this project's walls. The opening and fill survey (placement parent, body representation of doors, windows and openings, extrusion direction of opening profiles) used the seven development projects only, and none of the 13 test projects.

\section{The distillation corpus and the reinforcement-learning pool}
\label{sec:supp_data}

\paragraph{Size and provenance}
The supervised corpus holds 19,141 training episodes on 3,059 synthetic buildings and 2,270 validation episodes on 340 other buildings; no building appears in both splits, and no real project appears in either. Every episode was produced by Kimi K3 in thinking mode with the edit recipe of the main text and no checker feedback during the episode. Of the 24,276 teacher episodes that were run, 23,266 (95.8\%) passed all four checkers. Removing duplicate task solutions and episodes longer than the sequence limit (fewer than sixty) leaves 21,411 episodes across the training and validation splits. Of the 21,099 task prompts with a paraphrase record, 20,612 (97.7\%) were paraphrased; the remainder kept the template wording after the rewrite failed the guards twice. The 284 templates instantiate all task families of the main text (Table~\ref{tab:supp_corpus}): wall updates, including the three cascade tiers of wall moves, are the largest group, followed by retrieval, slab updates, quantity take-off, property edits and compound instructions. Each of the element kinds that IfcEditBench over-samples (doors, windows and openings) is represented by several hundred episodes per operation.

\begin{table}[!htbp]
 \centering
 \footnotesize
 \setlength{\tabcolsep}{4pt}
 \renewcommand{\arraystretch}{1.12}
 \caption{Composition of the training split of the distillation corpus (episodes) by target element and operation. Property edits, quantity take-off, retrieval and compound instructions are listed as their own rows; ``other'' collects the few updates of elements outside the main kinds.}
 \label{tab:supp_corpus}
 \begin{tabular}{@{}lrrrrr@{}}
 \toprule
 Target element & Create & Update & Delete & Retrieve & Total \\
 \midrule
 Wall & 336 & 4,581 & 296 & -- & 5,213 \\
 Slab & 663 & 1,204 & 382 & -- & 2,249 \\
 Column & 503 & 609 & 452 & -- & 1,564 \\
 Door & 530 & 253 & 740 & -- & 1,523 \\
 Window & 551 & 755 & 625 & -- & 1,931 \\
 Opening & -- & 216 & 206 & -- & 422 \\
 Space & 384 & 781 & 349 & -- & 1,514 \\
 Property set & -- & 1,074 & -- & -- & 1,074 \\
 Quantity take-off & -- & -- & -- & 1,118 & 1,118 \\
 Retrieval (identifiers, geometry, topology) & -- & -- & -- & 1,424 & 1,424 \\
 Compound instructions & \multicolumn{4}{c}{two operations each} & 1,043 \\
 Other & -- & 66 & -- & -- & 66 \\
 \midrule
 Total & 2,967 & 9,539 & 3,050 & 2,542 & 19,141 \\
 \bottomrule
 \end{tabular}
\end{table}

\paragraph{Tiers and reference styles}
By dependency tier the training split holds 3,536 episodes at tier 0, 13,205 at tier 1 and 1,200 each at tiers 2 and 3; the two deepest tiers are wall moves whose neighbouring walls or bounding spaces must follow. Among the grid-generator templates that vary the way an instruction refers to its target, 6,308 episodes name the target directly by identifier, 2,534 through a relationship, 2,450 through relative geometry, 645 along a path through the building and 111 from a viewpoint; the remaining 7,093 episodes come from the cascade, compound and single-form templates.

\paragraph{Buildings}
The 3,059 training buildings cover the export styles of the generator: 2,045 are modelled in metres and 1,014 in millimetres; 736 are exported as IFC2X3, 1,321 as IFC4 and 1,002 as IFC4X3; 602 carry the sparse relation set of a Revit-like export; 338 tile several plans along a corridor. Wall and space bodies are plain extrusions in 1,424 buildings, faceted Breps in 595, face sets in 649 and clipped extrusions in 391. Of the 3,000 buildings of the main generator, 1,090 have one storey, 825 two, 453 three, 452 four and the 180 remaining tiled buildings five to eight; 3,399 of the 3,400 generated buildings appear in the two splits. The 400 buildings of the second opening and fill generator (through-wall opening extrusions, rotated and polyline profiles, mapped door and window geometry) contribute 1,160 training and 129 validation episodes on door, window, opening and quantity tasks.

\paragraph{Episodes}
An episode is the message sequence of system prompt, task instruction, the teacher's reasoning and tool calls and the sandbox returns. The training episodes have a median of five assistant turns (mean 5.9, 90th percentile 8, maximum 31) and five tool calls (mean 6.2, 90th percentile 10, maximum 39); 959 episodes use at most two calls, 9,119 three to five, 7,553 six to ten and 1,510 more than ten. 15,440 episodes end with exactly one \texttt{commit()}, 1,128 commit two or more times, and 2,573 retrieval and quantity episodes answer without a commit. Measured with the student's tokenizer, an episode has a median length of 5,972 tokens (mean 7,216, 90th percentile 12,809, 99th percentile 22,901, maximum 31,278), of which the generated part, the reasoning, the tool calls and the final answer, is a median of 1,766 tokens (mean 2,646, 99th percentile 14,680), 37\% of all tokens; the sandbox returns and the fixed system prompt make up the rest. The teacher's reasoning is 42\% of the generated text. The validation split has the same distributions.

\paragraph{The reinforcement-learning pool}
The GRPO pool holds 5,895 tasks on 2,456 synthetic buildings: 1,095 at tier 0, 4,428 at tier 1, 165 at tier 2 and 207 at tier 3; by family, 1,829 creation, 1,302 update, 578 deletion, 849 retrieval and quantity, 426 compound, 664 cascade wall-move, 246 reference-resolution tasks and one spatial-reference task. 3,860 of them are tasks of the supervised corpus and 2,035 were generated for the pool and never shown to the teacher; 5,817 carry paraphrased prompts. Every task was attempted four times by the SFT student at temperature one before training: 458 were never solved, 689 once, 1,408 twice and 3,340 three times; tasks solved on all four attempts were discarded. This selection favours tasks that produce mixed outcomes in a group of eight rollouts.

\clearpage
\section{Task specification and the four checkers}
\label{sec:supp_checkers}

All tolerances are in metres; model units are converted with the file's unit scale (IfcOpenShell \texttt{calculate\_unit\_scale}), and meshes are generated in world coordinates with opening subtraction disabled.

\subsection{The specification}
A task record holds the identifier, project, template, dependency tier, instruction and input file, and the specification $\sigma$ in four fields: \texttt{target}, the GlobalIds the instruction is about (for a creation task the storey the new element goes to); \texttt{allowed}, the GlobalIds that may change without counting as side effects; \texttt{intent}, the predicate of the intent checker with its parameters; and \texttt{dependents}, a list of (GlobalId, rule) pairs for the dependency checker, with rule-specific parameters such as the move vector or the class of an expected new element. Creation tasks add \texttt{allowed\_new\_classes}, the classes whose new instances are expected. The first example is a tier-2 wall move in a Revit 2011 export; the two connected walls are allowed to change but carry no joint rule, because tier two relaxes joints. The second is a door creation.

{\footnotesize
\begin{verbatim}
{"task_id": "dental_clinic-T2-move-0122-s2", "project": "dental_clinic",
 "template": "T2-translate-wall-bounding-spaces", "tier": 2,
 "prompt": "Move the wall with GlobalId 3NhaIUfh12PAdrGa$S3xpR by 0.75 m in the -x direction of
   the project coordinate system. Its openings and the doors/windows in them must move with it and
   stay hosted, and the spaces it bounds and their space boundaries must follow the new wall face.
   Walls connected to it may be adjusted as needed. Do not change anything else.",
 "target": ["3NhaIUfh12PAdrGa$S3xpR"],
 "allowed": ["3NhaIUfh12PAdrGa$S3xpR", "3eM8WbY_59RR5TDWs3xrFu", "0ClPCUC7jCQRnj1dhupMRJ",
             "3Egy$LBd99weR14rW4bge1", "3NhaIUfh12PAdrGa$S3xqK"],
 "intent": {"type": "translate", "guid": "3NhaIUfh12PAdrGa$S3xpR", "vector": [-0.75, 0.0, 0.0], "tol": 0.02},
 "dependents": [
   {"guid": "3eM8WbY_59RR5TDWs3xrFu", "rule": "translate_with", "vector": [-0.75, 0.0, 0.0]},
   {"guid": "3eM8WbY_59RR5TDWs3xrFu", "rule": "stays_hosted"},
   {"guid": "0ClPCUC7jCQRnj1dhupMRJ", "rule": "space_meets_wall", "wall": "3NhaIUfh12PAdrGa$S3xpR",
    "vector": [-0.75, 0.0, 0.0]},
   {"guid": "0ClPCUC7jCQRnj1dhupMRJ", "rule": "boundary_on_wall", "wall": "3NhaIUfh12PAdrGa$S3xpR",
    "vector": [-0.75, 0.0, 0.0]}],
 "meta": {"n_openings": 1, "n_spaces": 1, "n_connected": 2}, "ifc": ".../dental_clinic/arc.ifc"}

{"task_id": "4351-C-door-0174-s2", "project": "4351", "template": "C-create-door", "tier": 1,
 "prompt": "Add a door 0.9 m wide and 2.1 m high to the wall with GlobalId 0S4TGKDX5FsvETM2lopwXc,
   anywhere along the wall. The door must be hosted by the wall through an opening
   (IfcRelVoidsElement / IfcRelFillsElement). Do not change anything else.",
 "target": ["0S4TGKDX5FsvETM2lopwXc"], "allowed": ["0S4TGKDX5FsvETM2lopwXc"],
 "allowed_new_classes": ["IfcOpeningElement", "IfcDoor"],
 "intent": {"type": "create_fill", "host": "0S4TGKDX5FsvETM2lopwXc", "kind": "door",
            "width": 0.9, "height": 2.1, "tol": 0.03},
 "dependents": [], "ifc": ".../4351/arc.ifc"}
\end{verbatim}}

\subsection{Validity}
The checker opens $\mathcal{M}'$ (an unparsable file fails every gate) and applies three tests. (i) A cycle in the decomposition graph: following \texttt{IfcRelAggregates} and \texttt{IfcRelNests} from every part, no object may reach itself. The test runs first because the geometry kernel does not terminate on such a model. A prompted model produced this failure by aggregating a storey into itself. (ii) New schema errors: the IfcOpenShell validator is run on $\mathcal{M}$ and on $\mathcal{M}'$ without EXPRESS rules, which checks that no abstract entity is instantiated and that every attribute value has the declared type, enumeration member, aggregate element type and cardinality, including inverse cardinalities; each logged line is one error, and the gate fails if $\mathcal{M}'$ has an error line that $\mathcal{M}$ does not. (iii) Geometry: for every product the edit touched (added or modified, Section~\ref{sec:supp_preservation}) that has a representation, the IfcOpenShell geometry kernel must build its shape without raising. The gate does not evaluate EXPRESS where-rules, geometric plausibility such as clashes or overlaps, the geometry of untouched products, or anything already wrong in the input export.

\subsection{Intent}
Table~\ref{tab:supp_intent} lists the intent types with their test and tolerance. Displacements are measured on mesh centroids in world coordinates, falling back to the placement origin when a mesh is unavailable. The task files set the tolerances explicitly; the table gives the values used in IfcEditBench. A compound intent is a list of parts that must all pass. A retrieval intent is judged on the last JSON object of the final message that carries a \texttt{value}, \texttt{global\_id} or \texttt{global\_ids} key. An intent type unknown to the checker fails.

\begin{table}[!htbp]
 \centering
 \footnotesize
 \setlength{\tabcolsep}{4pt}
 \renewcommand{\arraystretch}{1.1}
 \caption{Intent types: what is tested on the target products of $\mathcal{M}'$ and the tolerance used in IfcEditBench.}
 \label{tab:supp_intent}
 \begin{tabular}{@{}p{0.2\textwidth}p{0.62\textwidth}p{0.14\textwidth}@{}}
 \toprule
 Intent & Test & Tolerance \\
 \midrule
 translate & centroid displacement equals the requested vector & 0.02 m \\
 set\_attribute, set\_property & the attribute, or the property in the named property set, equals the value (numbers: relative tolerance) & exact; 0 \\
 delete, delete\_all & the product (every listed product) no longer exists & -- \\
 resize\_height & mesh height equals the requested height; where given, the top or bottom elevation as well & 0.02 m \\
 wall\_axis (shorten, rotate) & axis end points match the requested points in either order, and the body follows the axis: both ends moved like the nearest axis end (parallel case) or the body's principal direction aligned with the new axis with its length kept & 0.03 m per end; 0.1 m body \\
 set\_thickness, set\_thickness\_side & profile and body thickness across the axis; for the one-sided variant the fixed face stays in place & 0.02 m \\
 bbox, create\_wall\_bbox & bounding box of the (new) product equals the requested box & 0.05 m \\
 create\_fill & exactly one new door or window, hosted through an opening that voids the host wall, with the requested width and height, within the wall's length, not overlapping another opening; sill and plan centre where given & 0.03 m \\
 create\_wall, create\_column & new product with the requested axis end points (sum of end errors), thickness and height, or plan centre, size and height; contained in the requested storey & 0.03 m \\
 create\_slab, create\_space, space\_footprint & footprint polygon within 5\,\% of the requested area in symmetric difference (3\,\% for an edited space footprint) plus 0.05--0.1 m$^2$; thickness, height or top elevation; storey & 0.03--0.1 m \\
 rehost\_fill, opening\_width, create\_slab\_opening & the fill's opening voids exactly the new host at the requested position; the opening's extent along the wall and its centre; a new opening in a slab with the requested plan box through the slab's thickness & 0.02--0.05 m \\
 answer & identifier: exact; identifier set: equal; number: within the larger of an absolute and a relative tolerance (areas 1\,\%, quantities 2\,\%, counts exact) & as stated \\
 \bottomrule
 \end{tabular}
\end{table}

\subsection{Dependency}
Every dependent carries one rule; the gate is the conjunction over all dependents, and a rule the checker does not know fails. Table~\ref{tab:supp_rules} lists the rules with their tests, and the number of dependents using each rule in IfcEditBench. Rules that compare with the input are vacuous when the input lacks the evidence: a space that did not meet the wall in $\mathcal{M}$, or a space boundary without evaluable connection geometry, passes its rule with the verdict ``not applicable''. Rules test outcomes, so an opening that moves with its wall through its relative placement passes \texttt{translate\_with} without an explicit edit.

\begin{table}[!htbp]
 \centering
 \footnotesize
 \setlength{\tabcolsep}{4pt}
 \renewcommand{\arraystretch}{1.1}
 \caption{Dependency rules: the test on the dependent in $\mathcal{M}'$, and the number of dependents using the rule across the 1,990 tasks of IfcEditBench.}
 \label{tab:supp_rules}
 \begin{tabular}{@{}p{0.19\textwidth}p{0.69\textwidth}r@{}}
 \toprule
 Rule & Test & Uses \\
 \midrule
 translate\_with & the dependent's centroid displacement equals the target's move vector (0.02 m) & 958 \\
 stays\_hosted & the opening still voids a wall and, if it had a fill in $\mathcal{M}$, still has one & 614 \\
 removed, present & the dependent no longer exists; still exists & 487, 32 \\
 space\_meets\_wall & along the move axis, the nearest face of the space mesh (restricted to the wall's extent) lies within 0.02 m of a face of the moved wall, as it did in $\mathcal{M}$ & 374 \\
 boundary\_on\_wall & every space boundary between the space and the wall with connection geometry, transformed through the space's placement, lies within 0.02 m of the wall's centre or faces along the move axis, as it did in $\mathcal{M}$ & 374 \\
 unchanged & the dependent is neither removed nor modified (no attribute, property, placement or geometry change) & 353 \\
 joined\_to & the surface gap between the dependent wall and the moved wall is no larger than in $\mathcal{M}$ plus 0.02 m; the gap is zero when the solids interpenetrate, else the minimum point-to-triangle distance between surface samples & 278 \\
 space\_boundary\_with, space\_boundary\_new, new\_space\_bounds & the space's boundaries include the element; a new element of the class has a boundary to the space; a new space has a boundary to the element & 183, 26, 13 \\
 new\_in\_storey, in\_storey & every new element of the class is contained in the storey; the element's only container is the storey & 137, 36 \\
 connected\_new, rel\_connects, not\_connected & a new element of the class is connected to the wall; the two elements are connected by a connection relationship; they are not & 89, 24, 10 \\
 \bottomrule
 \end{tabular}
\end{table}

\subsection{Preservation}
\label{sec:supp_preservation}
The diff compares two graphs built from $\mathcal{M}$ and $\mathcal{M}'$. A \emph{node} is an \texttt{IfcProduct} instance with a GlobalId: elements, openings, spaces, storeys, building and site. Its content is the class; every scalar direct attribute except GlobalId, OwnerHistory, Tag, Description and LongName; every property of its property sets, flattened as ``set.property'' (type-level sets inherited, quantity sets not included); a content hash of its representation subtree; and a content hash of its placement subtree. The hashes follow every referenced entity to depth 24 and ignore STEP identifiers and GlobalIds. Re-serialising equivalent geometry in a different structure therefore counts as a modification. The placement hash follows relative-placement chains, so products placed relative to a moved wall also count as modified. The generator includes openings and fills in the allowed set for this reason. An \emph{edge} is a relationship instance keyed by class and end points: voids (wall to opening), fills (opening to fill), containment (structure to element), aggregation (object to part), path and element connections, space boundaries (space to element; all subtypes under one class name), nesting and port connections, and type and material assignments (element to the assigned type or material name). Property definitions are folded into the node and are not edges. A node is \emph{modified} when its class, attributes, representation hash or placement hash differ; edges are added or removed by key.

With $\mathcal{A}$ as in the main text (target, allowed, listed dependents, and new instances of the expected classes), the gate fails on any added, removed or modified node outside $\mathcal{A}$ and on any added or removed edge whose two ends are both outside $\mathcal{A}$ (an assignment edge counts as having one end outside). An edge with one end in $\mathcal{A}$ is exempt from the preservation check. Removing the moved wall's voids relationship or its boundary to a listed space therefore passes preservation; the dependency rules \texttt{stays\_hosted} and \texttt{boundary\_on\_wall} detect those removals. Conversely, the removal of a space boundary between an unlisted space and an unlisted wall changes no node, since boundaries are not products, and is caught only by the edge condition. The gate applies no geometric tolerance; the tolerances of Tables~\ref{tab:supp_intent} and \ref{tab:supp_rules} belong to the other two checkers.

\subsection{Automatic and explicit dependent changes}
Table~\ref{tab:supp_auto} separates what follows from the model when the agent edits the target from what the agent must do. The helpers edit one product each; the only propagation they perform is the one IFC itself performs through relative placement, and the relationship clean-up of a deletion. Before any edit, a helper gives the product a private copy of a sub-entity it shares with another product (a placement, a profile, a point, a property set), so that an edit of one wall does not move or re-label another.

\begin{table}[!htbp]
 \centering
 \footnotesize
 \setlength{\tabcolsep}{4pt}
 \renewcommand{\arraystretch}{1.1}
 \caption{Dependent changes that follow from the model or from a helper, and those the agent must make explicitly; the right column names the checker rule that tests the outcome.}
 \label{tab:supp_auto}
 \begin{tabular}{@{}p{0.3\textwidth}p{0.46\textwidth}p{0.2\textwidth}@{}}
 \toprule
 Change & How it happens & Tested by \\
 \midrule
 openings and fills placed relative to the moved wall & follow through the placement chain when \texttt{translate} or \texttt{rotate} edits the wall's own placement; \texttt{inspect} reports whether an opening or fill is placed relative to its wall & translate\_with \\
 openings and fills with an absolute placement & the agent translates each & translate\_with \\
 relationships of a deleted product & \texttt{delete} removes the product with its voids, fills, containment, aggregation, connection, boundary, type, material and property relationships and repairs geometry shared with survivors; it does not delete the opening of a deleted door or the openings of a deleted wall & removed, stays\_hosted \\
 space footprint on the moved wall face & the agent moves the vertices on that face (\texttt{shift\_profile\_vertices}, \texttt{shift\_vertices\_on\_plane}) & space\_meets\_wall \\
 space-boundary surface on the moved wall & the agent moves the boundary curve (\texttt{shift\_boundary\_curve}); it is placed relative to the space and never follows the wall & boundary\_on\_wall \\
 joined wall ends & the agent lengthens or shortens the touching end along its own axis (\texttt{move\_wall\_end}, \texttt{stretch\_wall\_end}) & joined\_to \\
 containment, connections and boundaries of a new element & \texttt{add\_wall}, \texttt{add\_slab}, \texttt{add\_column} contain the element in the storey and \texttt{add\_space} aggregates the space; connections and boundaries are separate steps (\texttt{connect\_walls}, IfcOpenShell API) & new\_in\_storey, connected\_new, space\_boundary\_new \\
 \bottomrule
 \end{tabular}
\end{table}

\subsection{Exports without explicit relationships}
Both generators derive dependents from the relationships in the file and never infer one from geometry. The cascade generator reads the spaces a wall bounds from its space boundaries and the walls it is joined to from its path connections, and assigns the tier from what it finds: a wall with openings but without boundaries or connections yields a tier-1 move whose dependents are the openings and fills; each bounded space raises the task to tier two and adds the two space rules; each connected wall raises it to tier three and adds the joint rule. The grid generator indexes connections and boundaries the same way and uses geometry only to phrase references, to pick the two walls a new wall abuts, and to test touching ends; a template that needs a bounded space, such as a wall identified by its inward side or a space adjacency question, skips a wall without boundaries. Table~\ref{tab:supp_inventory_b} shows the consequence: deep-tier tasks exist only in the projects that export the relationships.

\subsection{Passing and failing edits}
Table~\ref{tab:supp_gate_examples} gives, for each gate, one edit that passes and one that fails, for a tier-three wall move with an opening, a door and a joined wall, and for the door creation of the second example above.

\begin{table}[!htbp]
 \centering
 \footnotesize
 \setlength{\tabcolsep}{4pt}
 \renewcommand{\arraystretch}{1.1}
 \caption{One passing and one failing edit per gate, on a wall move by $-0.25$ m in $x$ with an opening, a door and a joined wall, and on a door creation.}
 \label{tab:supp_gate_examples}
 \begin{tabular}{@{}p{0.11\textwidth}p{0.42\textwidth}p{0.43\textwidth}@{}}
 \toprule
 Gate & Passes & Fails \\
 \midrule
 Validity & the wall's placement is edited; the validator logs the same lines as before and the wall, its opening and door rebuild & a predefined type set to a value outside its enumeration (new validator line); a profile replaced by a two-point polyline (the wall no longer rebuilds); a storey aggregated into itself (cycle) \\
 Intent & the wall's centroid moved by $(-0.25, 0, 0)$ within 0.02 m; exactly one new door, hosted through an opening of the named wall, $0.90\times2.10$ m within 0.03 m, inside the wall's length & the wall moved by $+0.25$ m; two doors created; a property written to a set whose name carries a trailing space, so the lookup finds nothing \\
 Dependency & opening and door moved by the same vector and still hosted; the joined wall's end stretched so that its surface gap to the moved wall is no larger than before & the joined wall left in place (gap 0.25 m); the door deleted instead of moved; at tier two, the space footprint or boundary curve left 0.25 m from the new wall face \\
 Preservation & only the wall, its opening and door, the listed spaces and joined walls appear in the diff, and every changed relationship has an end among them & a profile shared with a second wall edited in place, so that wall's geometry hash changes; the storey placement moved, so every element counts as modified; a new space created in a wall-creation task; a containment relationship of an unrelated element removed \\
 \bottomrule
 \end{tabular}
\end{table}

\clearpage
\section{The helper package}
\label{sec:supp_tools}

The agent can call the helpers in \texttt{h} from \texttt{execute\_ifc\_code}. Table~\ref{tab:supp_tools} gives their arguments and effects.

\begin{table}[!htbp]
\caption{The helper package \texttt{h} with its arguments and effects, grouped by inspection, element-local editing and creation. Bracketed arguments are optional. Helpers act on the specified products; propagation to other dependents is selected by the agent.}
\label{tab:supp_tools}
\footnotesize
\setlength{\tabcolsep}{4pt}
\renewcommand{\arraystretch}{1.1}
\begin{tabular*}{\textwidth}{@{}p{0.24\textwidth}p{0.25\textwidth}p{\dimexpr0.51\textwidth-4\tabcolsep\relax}@{}}
\toprule
\textbf{Helper} & \textbf{Arguments} & \textbf{Returns or effect} \\
\midrule
\rowcolor{black!7}\multicolumn{3}{@{}l}{\textbf{Inspection}\quad\emph{read-only functions of the model}} \\
\texttt{inspect} & entity, [psets], [related\_to] & placement chain, representation summary, openings and fills, connected walls and connection types, space boundaries and bounded spaces, shared sub-entities, property sets \\
\texttt{find} & [class], [storey], [name], [near, radius], [type] & the matching elements \\
\texttt{unit\_scale}, \texttt{world\_matrix}, \texttt{wall\_axis\_world}, \texttt{profile\_world\_polygon} & model or product & metres per model unit; world placement; wall axis and footprint profile in world coordinates \\
\texttt{wall\_length}, \texttt{space\_area} & wall or space & axis length; net floor area \\
\texttt{check\_model} & path of the input model & new schema errors; entities added, removed and modified against the input \\
\addlinespace[2pt]
\rowcolor{black!7}\multicolumn{3}{@{}l}{\textbf{Element-local edits}\quad\emph{one product each; shared sub-entities are copied before they are changed}} \\
\texttt{translate} & product, vector & moves one placement; children placed relative to it follow \\
\texttt{move\_wall\_end}, \texttt{stretch\_wall\_end} & wall, end, vector or delta & moves or lengthens one end of a wall along its axis \\
\texttt{shift\_vertices\_on\_plane}, \texttt{shift\_profile\_vertices} & product, plane, vector, [within] & moves every vertex of the body (extrusion, Brep, polyline) or of the footprint profile that lies on a world plane \\
\texttt{shift\_boundary\_curve} & space boundary, [plane], vector & moves a space-boundary surface \\
\texttt{set\_extrusion\_depth}, \texttt{set\_wall\_thickness}, \texttt{rotate} & product, value & height; thickness about the axis; rotation about a world point \\
\texttt{set\_property}, \texttt{set\_attribute} & product, name, value & writes a private property set or an attribute \\
\texttt{delete}, \texttt{connect\_walls}, \texttt{private} & product(s) & removes a product with its relationships cleaned; joins two walls; makes a shared sub-entity private \\
\addlinespace[2pt]
\rowcolor{black!7}\multicolumn{3}{@{}l}{\textbf{Creation}\quad\emph{new products on a storey or in a wall; the storey relationship is chosen by class}} \\
\texttt{add\_wall}, \texttt{add\_slab}, \texttt{add\_column} & storey, geometry & a new element with its own placement and body, contained in the storey through \texttt{IfcRelContainedInSpatialStructure} \\
\texttt{add\_space} & storey, footprint, height & a new \texttt{IfcSpace} aggregated into the storey through \texttt{IfcRelAggregates}, as the spatial structure prescribes \\
\texttt{add\_opening\_fill} & wall, position, size, kind & an opening and its door or window hosted by the wall \\
\bottomrule
\end{tabular*}
\end{table}

\section{Reproduction of training and evaluation}
\label{sec:supp_repro}

\paragraph{Software and hardware}
The sandbox, the checkers, the task generators and the BIM-Edit scorer run on Python 3.12 with IfcOpenShell 0.8.5 \cite{ifcopenshell} and NumPy 2.5; the same interpreter image executes the agent's code in every episode, so the helper package and the checkers see one IfcOpenShell version throughout. Training uses verl 0.9.0 \cite{verl2024}, whose hybrid engine serves rollouts with vLLM 0.24 \cite{vllm2023}; the students are evaluated with the same vLLM build in bfloat16. DeepSeek-V4-Flash is served with the vLLM build released for that model and Kimi K3 with SGLang \cite{sglang2024}, both at a 65,536-token context window (Section~\ref{sec:supp_runs} lists the per-run settings). Model weights are the public releases of Qwen3.5-9B, Qwen3.5-27B, Gemma 4 31B, DeepSeek-V4-Flash and Kimi K3. All runs used nodes with eight H100 80\,GB GPUs: SFT on one node, 9B GRPO on four nodes, the intent-and-validity GRPO and the 27B GRPO on two nodes, Kimi K3 inference on four nodes.

\paragraph{Prompts}
The student's system prompt documents the single tool \texttt{execute\_ifc\_code} and the 28 helpers of the package \texttt{h} with their arguments and effects (Section~\ref{sec:supp_tools}); a budget note is appended to the tool return when two rounds remain. The teacher receives, in addition, an edit recipe of about 12,000 characters: step-by-step procedures for a wall move (axis end points, faces, connected walls, openings, bounded spaces and their boundaries), for re-hosting a fill, for deletions with relationship clean-up and for resolving spatial, topological and path references. The student learns from the resulting traces, with the recipe omitted from its prompt. Both prompts and the recipe are released verbatim with the code.

\paragraph{Determinism}
Evaluation decodes at temperature 0; Section~\ref{sec:supp_runs} lists the evaluation behind every row.

\paragraph{Training hyperparameters}
Both student sizes use the training settings in Table~\ref{tab:supp_hparams}. Settings are fixed independently of the test set. SFT uses the final checkpoint, and GRPO runs for a fixed number of steps with the task pool and online group filter described in the main text. The adapters are merged into the rollout weights after every step so that the sampling policy and the trained policy coincide.

\begin{table}[!htbp]
 \centering
 \footnotesize
 \setlength{\tabcolsep}{3pt}
 \renewcommand{\arraystretch}{1.12}
 \caption{Training hyperparameters of the two stages.}
 \label{tab:supp_hparams}
 \begin{tabular}{@{}lp{0.16\linewidth}p{0.72\linewidth}@{}}
 \toprule
 Stage & Setting & Value \\
 \midrule
 SFT & adapters & LoRA on all linear layers, rank 64, scaling 128 \\
 SFT & optimiser & learning rate $10^{-4}$, cosine decay, 3 epochs, batch size 16 \\
 SFT & sequence length & up to 32,768 tokens; final checkpoint used without selection \\
 GRPO & pool & synthetic tasks, prefiltered offline with 4 attempts of the SFT student at temperature 1; unsolved tasks capped at one fifth \\
 GRPO & sampling & 64 prompts per step, $G=8$ rollouts each, evaluation protocol (20 rounds, no checker feedback) \\
 GRPO & filtering & groups with eight equal rewards dropped and resampled until 64 mixed groups \\
 GRPO & optimiser & mini-batches of 8 prompts, learning rate $10^{-5}$, KL coefficient $\beta=10^{-2}$ towards the SFT policy \\
 GRPO & schedule & 100 steps, adapters merged into the rollout weights after every step \\
 \bottomrule
 \end{tabular}
\end{table}

\section{Evaluation runs and run-to-run variation}
\label{sec:supp_runs}

Four evaluations supply the reported IfcEditBench scores: I, the students and the untrained 27B model of the main table; II, the prompted and framework references; III, the ablations and the untrained 9B model; IV, the 27B GRPO checkpoints; the 9B GRPO checkpoints have their own evaluations. Table~\ref{tab:supp_runs} identifies the weights and evaluation behind each reported score. We use one IfcEditBench score per model throughout the paper and report its external-benchmark scores from the same evaluation. All evaluations use the protocol of the main text (temperature 0, 16,384 output tokens per turn, budget 20, or 25 on BIBIMBAP, no checker feedback) unless the table says otherwise. A task without an outcome counts as a failure in the overall score and is dropped from paired comparisons.

\begin{table}[!htbp]
 \centering
 \footnotesize
 \setlength{\tabcolsep}{3pt}
 \renewcommand{\arraystretch}{1.1}
 \caption{Correspondence of the reported IfcEditBench rows to weights and evaluations (I--IV as defined in the text; every evaluation covers the 1,500 held-out and the 490 fill tasks), with each model's IfcEditBench strict success (\%).}
 \label{tab:supp_runs}
 \begin{tabular}{@{}p{0.24\textwidth}p{0.23\textwidth}p{0.14\textwidth}p{0.19\textwidth}r@{}}
 \toprule
 Row & Weights & Evaluation & Settings beyond the protocol & Strict \\
 \midrule
 Qwen3.5-9B (untrained) & public Qwen3.5-9B & III & -- & 16.08 \\
 Qwen3.5-27B (untrained) & public Qwen3.5-27B & I & -- & 59.35 \\
 Qwen3.5-9B, SFT & SFT, final checkpoint (3,588 steps) & I & -- & 79.15 \\
 Qwen3.5-9B, SFT + GRPO & GRPO step 100 from the SFT weights & I & -- & 86.78 \\
 Qwen3.5-27B, SFT & SFT, final checkpoint & I & -- & 88.29 \\
 Qwen3.5-27B, SFT + GRPO & GRPO step 100 from the 27B SFT weights & IV & -- & 91.76 \\
 DeepSeek-V4-Flash & public weights, vLLM, tensor parallel 4 & II & 65,536-token context & 61.11 \\
 Gemma 4 31B & public weights, SGLang & II & 131,072-token context & 67.99 \\
 Kimi K3 & public weights, SGLang on four nodes & II & 65,536-token context, reasoning effort high & 88.04 \\
 MCP4IFC & DeepSeek-V4-Flash backbone, 39 tools & II & tool output capped at 8,000 characters & 39.40 \\
 MCP4IFC, looser cap & same & II & cap 60,000 characters & 33.42 \\
 IFC-Agent & DeepSeek-V4-Flash backbone & II & own execution loop & 13.62 \\
 \midrule
 \multicolumn{5}{@{}l}{\emph{Ablations and curves}} \\
 reward on intent and validity only & GRPO step 100 with the intent-and-validity reward & III & -- & 82.01 \\
 SFT on 25 / 50 / 75\,\% of the corpus & SFT on nested subsets & III & -- & 75.38, 78.39, 79.35 \\
 final student, budget 40 & as the final student & III; fill tasks separate & budget 40 & 86.38 \\
 final student with verifier & as the final student & III; fill tasks separate & checker feedback after every step & 87.29 \\
 9B GRPO steps 25, 50, 75 & exported checkpoints of the 9B GRPO run & separate & -- & 83.57, 85.43, 85.93 \\
 27B GRPO steps 10--75 & exported checkpoints of the 27B GRPO run & IV & -- & 90.15--92.01 \\
 \bottomrule
 \end{tabular}
\end{table}

\paragraph{Training runs}
Table~\ref{tab:supp_trainruns} lists the training runs behind the reported students. The 9B GRPO run was restarted once from its step-65 checkpoint with a fresh optimiser state; the 27B run was not interrupted.

\begin{table}[!htbp]
 \centering
 \footnotesize
 \setlength{\tabcolsep}{3pt}
 \renewcommand{\arraystretch}{1.1}
 \caption{Training runs: starting weights, data, compute and the checkpoint that was served.}
 \label{tab:supp_trainruns}
 \begin{tabular}{@{}p{0.2\textwidth}p{0.17\textwidth}p{0.24\textwidth}p{0.16\textwidth}p{0.19\textwidth}@{}}
 \toprule
 Run & Start & Data & Compute & Served checkpoint \\
 \midrule
 9B SFT & Qwen3.5-9B & 19,141 episodes (3 epochs, 3,588 steps) & 8 H100 & final step, adapters merged \\
 27B SFT & Qwen3.5-27B & same & 8 H100 & final step, merged \\
 9B SFT on 25 / 50 / 75\,\% & Qwen3.5-9B & nested subsets of 4,785 / 9,570 / 14,356 episodes & 8 H100 & final step, merged \\
 9B GRPO (final student) & 9B SFT & pool of 5,895 tasks & 32 H100, 18 h; restarted once from step 65 & step 100, merged; steps 25, 50, 75 exported \\
 9B GRPO, intent-and-validity reward & 9B SFT & same pool & 16 H100, 23 h & step 100, merged \\
 27B GRPO & 27B SFT & same pool & 16 H100, 49 h & step 100, merged; steps 10--75 exported \\
 \bottomrule
 \end{tabular}
\end{table}

\section{Reproduction of the BIM-Edit protocol and scorer}
\label{sec:scorer}

The official BIM-Edit evaluator is not public. We re-implemented it from the benchmark paper and the evaluator defaults shipped with the released runs, and fitted the choices the paper leaves open to released per-task scores before any model of ours was scored.

\paragraph{Scoring formula}
For a task with input model $\mathcal{M}$, ground truth $\mathcal{M}^{\star}$ and prediction $\mathcal{M}'$, the scorer builds the two edit graphs $\mathrm{diff}(\mathcal{M},\mathcal{M}^{\star})$ and $\mathrm{diff}(\mathcal{M},\mathcal{M}')$ with the graph of Section~\ref{sec:supp_checkers} (products as nodes, relationship instances as edges). The \emph{live} nodes of each side are its added and modified products; each receives an area-weighted surface point cloud of $\max(64,\min(4096,\lfloor 16384/n\rfloor))$ points for $n$ live nodes. Live nodes of the two sides are matched by Hungarian assignment on the Monte-Carlo intersection over union of their oriented bounding boxes, with identical GlobalIds matched at once, pairs of different IFC classes excluded, and a pair kept only above an overlap of 0.05. For creation and update tasks, \emph{geometry} is $\exp(-5\,d/D)$, where $d$ is the symmetric median nearest-neighbour distance between the pooled ground-truth and pooled prediction clouds and $D$ the diagonal of their joint bounding box; \emph{semantics} averages over the ground-truth live nodes one half for a class match plus one half times the share of the node's attributes and property values (tolerance 5\,\% on numbers) that the matched prediction node reproduces, with unmatched nodes scoring zero; \emph{topology} is $0.3\,F_{1}^{\mathrm{node}}+0.7\,F_{1}^{\mathrm{rel}}$, the $F_1$ scores of the edited-node sets and of the added and removed relationship tuples over six relationship classes (voids, fills, containment, aggregation, path connection, space boundary). For deletion tasks, geometry is one when every ground-truth-removed product is absent from the prediction and zero otherwise, and semantics takes the complementary value, the choice that fitted the released scores best. The final score is the mean of the three axes; an unparsable prediction scores zero on all of them, and on anchored tasks a prediction that drops the pinned target scores zero.

\paragraph{Fit}
Eight choices were left open by the benchmark description: the alignment of the two sides (edit graphs or GlobalIds), the attribute set for semantics (all or task-relevant), the box type for matching (oriented or axis-aligned), class-matched pairs only or not, zero score without any match or not, the delete-semantics convention, the overlap threshold (0.05 or 0.2) and the relationship-class set (four, six, eight or eleven classes), together with the anchored pin; all 1,024 combinations were enumerated. The fixed constants (chamfer scale 5, node weight 0.3, tolerance 0.05, sample counts) are the released evaluator defaults. The objective was the mean absolute difference between our and the official final score over the 324 tasks, averaged over two released runs with edited files, Claude Sonnet 4.6 and Gemini 3.0 Flash; the two other released runs with edited files, GPT-5.4 Pro and GPT-5.4 Mini, were held out and used only to report the error. The best configuration (edit-graph alignment, all attributes, oriented boxes, class-matched pairs, no zeroing, present-is-one delete semantics, threshold 0.05, six relationship classes, pin on) reaches a mean absolute error of 20.8 points on the calibration runs and is the configuration used for every BIM-Edit score in this paper. Table~\ref{tab:scorer_fit} gives the error per run: on the two held-out runs the per-task mean absolute error on the final score is 20 points, the share of tasks within 5 points is 35--41\,\%, and the run mean is reproduced within 2.2 points.

\begin{table}[!htbp]
 \centering
 \footnotesize
 \setlength{\tabcolsep}{3pt}
 \renewcommand{\arraystretch}{1.12}
 \caption{Agreement of the re-implemented scorer with the official per-task scores on the four released runs with edited files (324 tasks each; points on a 0--100 scale). Calibration runs were used to select the configuration; held-out runs were not. Within 5 = share of tasks whose final score agrees within 5 points; bias = ours minus official on the run mean.}
 \label{tab:scorer_fit}
 \begin{tabular}{@{}llccccccc@{}}
 \toprule
 Run & Role & MAE final & MAE geom. & MAE sem. & MAE topo. & Within 5 & Mean ours / official & Bias \\
 \midrule
 Claude Sonnet 4.6 & calibration & 18.4 & 26.2 & 27.0 & 20.4 & 0.46 & 42.06 / 45.31 & $-3.26$ \\
 Gemini 3.0 Flash & calibration & 23.2 & 27.6 & 33.9 & 38.4 & 0.20 & 50.48 / 49.48 & $+1.00$ \\
 GPT-5.4 Pro & held-out & 20.3 & 29.9 & 28.5 & 23.8 & 0.35 & 45.26 / 43.94 & $+1.32$ \\
 GPT-5.4 Mini & held-out & 20.1 & 27.5 & 27.4 & 21.5 & 0.41 & 37.57 / 39.78 & $-2.21$ \\
 \midrule
 All four & & 20.5 & 27.8 & 29.2 & 26.0 & 0.36 & 43.84 / 44.63 & $-0.79$ \\
 \bottomrule
 \end{tabular}
\end{table}

\paragraph{Per-operation bias}
Table~\ref{tab:scorer_ops} splits the agreement by operation over the four released runs. For creations and deletions, the final-score error is 9--14 points, the per-task correlation is 0.7--0.8 for creations and for deletions on three of the four runs, and the run means agree within 1.3 points. Updates have a per-task error of 39 points and a correlation of at most 0.24. Their semantics bias ($-22$ points) and topology bias ($+24$ points) partly cancel in the final score ($-2.3$ points). On deletions the geometry and semantics errors are mirror images ($-20$ and $+20$ points) and cancel exactly, since both axes flip together whenever a removed product survives. All our models use this scorer. It reproduces run means within about two points, but update scores per task and update/deletion scores per axis cannot be compared directly with published values.

\begin{table}[!htbp]
 \centering
 \footnotesize
 \setlength{\tabcolsep}{3pt}
 \renewcommand{\arraystretch}{1.12}
 \caption{Per-operation agreement with the official scores, pooled over the four released runs (432 tasks per operation; points). Bias = ours minus official on the mean; $r$ = per-task Pearson correlation of the final score (range over the four runs).}
 \label{tab:scorer_ops}
 \begin{tabular}{@{}lcccccccc@{}}
 \toprule
 & \multicolumn{4}{c}{Final score} & \multicolumn{3}{c}{Bias per axis} \\
 \cmidrule(lr){2-5}\cmidrule(lr){6-8}
 Operation & Official & Ours & MAE & $r$ & Geom. & Sem. & Topo. \\
 \midrule
 Create & 25.97 & 27.14 & 14.3 & 0.75--0.78 & $+13.3$ & $-8.8$ & $-1.0$ \\
 Delete & 50.83 & 49.59 & 8.6 & $-0.12$--0.79 & $-19.9$ & $+20.1$ & $-3.9$ \\
 Update & 57.09 & 54.80 & 38.6 & $-0.02$--0.24 & $-8.3$ & $-22.1$ & $+23.6$ \\
 \bottomrule
 \end{tabular}
\end{table}

\paragraph{The Gemma 4 31B reproduction}
Our own run of Gemma 4 31B under the BIM-Edit protocol (Table~\ref{tab:gemma}) has similar runtime behaviour to the published run but a lower final score. Execution, budget-exhaustion and crash rates differ by less than two points, and tool-call counts are similar. The benchmark releases per-operation scores for its Gemma run but not its edited files, so the two runs are compared by operation only, and this model is not part of the calibration. The published and our final scores are 0.58 against 0.86 for creations, 53.97 against 46.84 for deletions and 58.08 against 43.96 for updates. The differences follow the measured scorer bias: lower geometry and higher semantics on deletions, lower semantics and higher topology on updates. This comparison combines scorer bias with differences between the two runs. The operation ordering and episode-completion rates agree with the published run.

\begin{table}[!htbp]
 \centering
 \footnotesize
 \setlength{\tabcolsep}{3pt}
 \caption{Gemma 4 31B, prompted CodeAct, under the BIM-Edit protocol: our harness and re-implemented scorer against the published row; 95\% bootstrap interval on our final score.}
 \label{tab:gemma}
 \begin{tabular}{@{}lcc@{}}
 \toprule
 Gemma 4 31B, prompted CodeAct & Ours & Published \\
 \midrule
 Final score & 30.55 $[27.27, 33.31]$ & 37.54 \\
 Create & 0.9 & 0.6 \\
 Delete & 46.8 & 54.0 \\
 Update & 44.0 & 58.1 \\
 Episodes executed (\%) & 84.0 & 85.19 \\
 Budget exhausted (\%) & 10.2 & 8.3 \\
 Crashed (\%) & 5.9 & 4.6 \\
 Tool calls per task & 9.57 & 8.57 \\
 \bottomrule
 \end{tabular}
\end{table}

\section{Results by dependency tier and by target element}
\label{sec:supp_tiers}

Tier 1 accounts for most of the test set (1,509 of 1,990 tasks) and therefore largely determines the overall score (Table~\ref{tab:supp_tiers}); tiers 2 and 3 are the 200 wall moves whose bounding spaces or joined walls must follow. The prompted models lose most of their success between tier 0 and tier 2: DeepSeek-V4-Flash falls from 85.8 to 13.0, Qwen3.5-27B from 76.5 to 25.0 and Gemma 4 31B from 80.4 to 38.0, and the two published frameworks reach 2.0 and 0.0 at tier 2. Kimi K3, the teacher, declines more gently (92.5 to 74.0). The trained students maintain higher success on these dependent edits: the 9B student after SFT scores 75.0 and 80.0 at tiers 2 and 3, after GRPO 82.0 and 89.0, and the 27B students 93.0 and 96.0 after SFT and 100.0 and 99.0 after GRPO. Training produces its largest advantage over prompting on the deeper dependency tiers.

\begin{table}[!htbp]
 \centering
 \footnotesize
 \setlength{\tabcolsep}{3pt}
 \renewcommand{\arraystretch}{1.12}
 \caption{Strict success (\%) on IfcEditBench by dependency tier for prompted models, frameworks and trained students (tier sizes 281, 1,509, 100, 100).}
 \label{tab:supp_tiers}
 \begin{tabular}{lccccc}
 \toprule
 Model & Tier 0 & Tier 1 & Tier 2 & Tier 3 & All \\
 \midrule
 DeepSeek-V4-Flash & 85.8 & 62.0 & 13.0 & 27.0 & 61.11 \\
 Gemma 4 31B & 80.4 & 68.2 & 38.0 & 60.0 & 67.99 \\
 Kimi K3 & 92.5 & 89.3 & 74.0 & 70.0 & 88.04 \\
 Qwen3.5-9B & 42.0 & 12.5 & 1.0 & 13.0 & 16.08 \\
 Qwen3.5-27B & 76.5 & 60.6 & 25.0 & 27.0 & 59.35 \\
 MCP4IFC & 75.4 & 36.8 & 2.0 & 15.0 & 39.40 \\
 IFC-Agent & 56.6 & 7.3 & 0.0 & 2.0 & 13.62 \\
 \midrule
 Qwen3.5-9B, SFT & 86.5 & 78.0 & 75.0 & 80.0 & 79.15 \\
 Qwen3.5-9B, SFT + GRPO & 89.0 & 86.5 & 82.0 & 89.0 & 86.78 \\
 Qwen3.5-27B, SFT & 91.1 & 86.9 & 93.0 & 96.0 & 88.29 \\
 Qwen3.5-27B, SFT + GRPO & 93.6 & 90.4 & 100.0 & 99.0 & 91.76 \\
 \bottomrule
 \end{tabular}
\end{table}

Table~\ref{tab:supp_pairs} gives paired differences by target-element group with their 95\% cluster-bootstrap intervals. Identifier retrieval and compound tasks contribute to the overall group only, as in the main-text figure. The two scale and training effects are of similar size overall ($+9.15$ for the larger student, $+7.75$ for GRPO on the 9B student) and both are largest on walls, slabs, columns and spaces; on property and quantity tasks neither interval excludes zero, and this group is where the 9B GRPO student trails the 27B SFT student most clearly ($+4.71$ [0.59, 8.88]). Between those two models, the interval for the overall difference of $+1.51$ includes zero. On door, window and opening tasks the point estimate is slightly negative and its interval also includes zero. The 27B GRPO student exceeds its SFT start by 3.42 points overall, by 4.73 on walls, slabs and columns, 2.98 on doors, windows and openings and 1.94 on spaces, and by 0.00 on property and quantity tasks.

\begin{table}[!htbp]
 \centering
 \footnotesize
 \setlength{\tabcolsep}{3pt}
 \renewcommand{\arraystretch}{1.12}
 \caption{Paired differences in strict success (points) with 95\% cluster-bootstrap intervals and the number of paired tasks, per target-element group.}
 \label{tab:supp_pairs}
 \begin{tabular}{@{}>{\raggedright\arraybackslash}p{0.16\linewidth}*{4}{C{0.195\linewidth}}@{}}
 \toprule
 Group & 27B SFT $-$ 9B SFT & 9B SFT + GRPO $-$ 9B SFT & 27B SFT $-$ 9B SFT + GRPO & 27B SFT + GRPO $-$ 27B SFT \\
 \midrule
 All tasks & $+9.15$ [7.42, 11.05] (1989) & $+7.75$ [5.79, 9.47] (1988) & $+1.51$ [-0.05, 3.48] (1987) & $+3.42$ $[1.99, 4.71]$ (1,989) \\
 Wall, slab, column & $+10.81$ [7.68, 13.92] (740) & $+7.98$ [4.48, 11.04] (739) & $+2.84$ [0.13, 5.93] (739) & $+4.73$ $[2.72, 6.67]$ (740) \\
 Door, window, opening & $+8.30$ [5.38, 11.18] (771) & $+8.68$ [6.00, 11.46] (772) & $-0.26$ [-2.99, 2.41] (771) & $+2.98$ $[0.77, 5.33]$ (771) \\
 Space & $+10.68$ [3.06, 17.86] (103) & $+10.68$ [3.00, 18.00] (103) & $+0.00$ [-5.71, 5.88] (103) & $+1.94$ $[-5.45, 9.28]$ (103) \\
 Property, quantity & $+2.92$ [-1.16, 7.14] (171) & $-1.18$ [-5.68, 3.47] (170) & $+4.71$ [0.59, 8.88] (170) & $+0.00$ $[-2.99, 3.01]$ (171) \\
 \bottomrule
 \end{tabular}
\end{table}

\section{The 27B GRPO run at every exported checkpoint}
\label{sec:supp_curve}

Table~\ref{tab:supp_curve} lists every exported checkpoint of the 27B GRPO run together with the SFT start (step 0) and the final student (step 100), on IfcEditBench and on both public benchmarks; the main text plots steps 0, 25, 50, 75 and 100 on IfcEditBench only. On IfcEditBench most of the gain arrives in the first ten steps (88.29 to 90.15) and the curve then climbs slowly to 92.01 at step 75 before settling at 91.76; no checkpoint falls below the SFT start. BIM-Edit moves from 54.76 to between 56.94 and 59.51 from step 10 onwards without a clear trend across checkpoints. BIBIMBAP, with 100 tasks, varies between 57.25 and 65.83 from checkpoint to checkpoint; this spread is comparable to the differences between models in the main-text table, limiting the precision of rankings based on their point estimates.

\begin{table}[!htbp]
 \centering
 \footnotesize
 \setlength{\tabcolsep}{3pt}
 \renewcommand{\arraystretch}{1.12}
 \caption{The 27B GRPO run: strict success (\%) on IfcEditBench, BIM-Edit final score and BIBIMBAP overall score at every exported checkpoint (step 0 = the SFT start, step 100 = the final 27B GRPO student).}
 \label{tab:supp_curve}
 \begin{tabular}{lccc}
 \toprule
 GRPO step & IfcEditBench & BIM-Edit & BIBIMBAP \\
 \midrule
 0 & 88.29 & 54.76 & 59.83 \\
 10 & 90.15 & 58.00 & 65.83 \\
 20 & 90.25 & 58.68 & 59.17 \\
 25 & 90.20 & 58.78 & 63.50 \\
 30 & 90.35 & 57.05 & 63.00 \\
 40 & 90.80 & 59.51 & 64.50 \\
 50 & 91.46 & 58.23 & 64.33 \\
 75 & 92.01 & 59.49 & 57.25 \\
 100 & 91.76 & 56.94 & 63.83 \\
 \bottomrule
 \end{tabular}
\end{table}

\section{Runtime outcomes}
\label{sec:supp_runtime}

Table~\ref{tab:supp_runtime} reports runtime outcomes for models evaluated with the shared CodeAct protocol. Executed denotes normal termination with a final answer, regardless of whether a commit was made or the checkers passed; budget and crash denote those termination conditions, and other collects the remaining outcomes. Budget exhaustion and under-editing are the dominant failure modes of the untrained open models: Qwen3.5-9B exhausts the budget on 44.6\% of the tasks and leaves 70.3\% under-edited, DeepSeek-V4-Flash and Qwen3.5-27B on about a fifth each, while crashes stay below five percent for every model. Training removes most of both: the 9B student after SFT exhausts the budget on 7.1\% and under-edits 7.9\%, after GRPO 5.2\% and 4.4\%, and the 27B students reach the execution rate of Gemma 4 31B and Kimi K3 (95.0--98.2\%) with fewer tool calls per task (6.86--7.58 against 7.37--9.10). The untrained models that fail also call tools more often (12--15 calls per task), consistent with repeated inspection without a commit. Under-edit is recorded separately when a non-retrieval task fails intent with no entities changed. Both under-edit and outcome rates use all tasks as their denominator.

\begin{table}[!htbp]
 \centering
 \footnotesize
 \setlength{\tabcolsep}{3pt}
 \renewcommand{\arraystretch}{1.12}
 \caption{Runtime outcomes (\%) on IfcEditBench and mean tool calls per task; the two published frameworks use their own execution loops and are omitted.}
 \label{tab:supp_runtime}
 \begin{tabular}{lcccccc}
 \toprule
 Model & Executed & Budget & Crash & Other & Under-edit & Calls / task \\
 \midrule
 DeepSeek-V4-Flash & 75.2 & 20.4 & 4.3 & 0.2 & 22.3 & 13.67 \\
 Gemma 4 31B & 95.5 & 4.2 & 0.3 & 0.1 & 2.7 & 7.37 \\
 Kimi K3 & 95.0 & 2.3 & 1.4 & 1.4 & 3.9 & 9.10 \\
 Qwen3.5-9B & 53.5 & 44.6 & 1.4 & 0.5 & 70.3 & 15.22 \\
 Qwen3.5-27B & 76.9 & 20.8 & 2.3 & 0.0 & 26.3 & 12.06 \\
 \midrule
 Qwen3.5-9B, SFT & 91.6 & 7.1 & 1.0 & 0.3 & 7.9 & 8.35 \\
 Qwen3.5-9B, SFT + GRPO & 94.5 & 5.2 & 0.2 & 0.2 & 4.4 & 8.07 \\
 Qwen3.5-27B, SFT & 96.1 & 2.9 & 0.8 & 0.3 & 2.9 & 7.58 \\
 Qwen3.5-27B, SFT + GRPO & 98.2 & 1.5 & 0.3 & 0.0 & 1.0 & 6.86 \\
 \bottomrule
 \end{tabular}
\end{table}

\section{Expert evaluation}
\label{sec:supp_humaneval}

\paragraph{Sampling}
The 30 IfcEditBench items are edits of the final 9B student from its main-text evaluation on the 1,500 held-out tasks, restricted to editing families (creation, deletion, update, compound and the cascade tiers; no retrieval or quantity task) and to tasks whose edited file exists. With a fixed random seed, 15 strict passes were drawn by a quota over task families (creations, deletions, updates, compound instructions, spatial references and the cascade tiers) and 15 failures by the first failing gate in the order validity, preservation, dependency, intent (one, three, five and six items), with at most one failure per project and template. The 20 benchmark pairs put the final student next to another system on the same task: on BIM-Edit against the released GPT-5.4 Pro outputs (five pairs) and the 9B SFT start (five), on BIBIMBAP against the 9B SFT start (five) and the 27B SFT student (five). Within each BIM-Edit group, pairs were drawn so that the benchmark score favours the final student in three, the other system in one, and is equal in one control pair, with at most one task per element--operation combination; within each BIBIMBAP group, four pairs differ by at least 0.25 with at most one favouring the other system, plus one control pair with equal scores; all our outputs come from the main-text evaluation.

\paragraph{Material and blinding}
The rater worked through a static web page built from the shuffled items. An IfcEditBench item shows the instruction, a ``before'' view (isometric and plan section, with the elements the instruction refers to dashed) and an ``after'' view (added elements green, removed red, modified orange), the list of changed products with class, name and storey, and the list of changed relationships. The rater answers whether the edited model is an acceptable execution of the instruction (yes, no, or cannot judge from this material), ticks reasons for a rejection (requested change missing or wrong, dependents not handled, side effects, invalid model, other), states a confidence (low, medium, high) and may comment. A pair item shows the input model and the two outputs as Output A and Output B in random order, with their change lists; the rater chooses the better output, a tie with both acceptable or a tie with both unacceptable, and rates each output's acceptability. The page carries no checker verdict, benchmark score or model name; the mapping from items to systems and verdicts was kept apart from it. The rater is one BIM practitioner, recorded under the label R07; no further personal information was collected.

\paragraph{Results on the 30 edits}
Table~\ref{tab:supp_he_items} lists every item. The rater judged 21 items and marked 9 as not judgeable, every one of them at low confidence; the recurring reason is that a verdict needed a number the views do not show (a slab's thickness or elevation, an opening's width) or a reference element that the section box does not identify (the nearest column, the lowest slab, the exterior wall). Of the 21 judged items, 19 agree with the four-gate verdict. In item A05, the checkers accept a door in the host wall named by the specification. The rater interprets ``the wall on your right'' as the opposite wall of the route and rejects that placement. In item A22 the checkers reject a tier-three wall move because the student also shifted two vertices of a basement space whose footprint abutted the moved wall; the export carries no space boundaries, so the generator had no relationship from which to list the space as a dependent, and the rater found the edit acceptable. A specification with that relationship would make the space a required dependent instead of a protected entity.

\begin{table}[!htbp]
 \centering
 \scriptsize
 \setlength{\tabcolsep}{2pt}
 \renewcommand{\arraystretch}{1.08}
 \caption{The 30 IfcEditBench items of the expert evaluation: task template and tier, the four-gate verdict with the failing gates, the rater's verdict and confidence, and the rater's reason, or the reason an item was not judgeable.}
 \label{tab:supp_he_items}
 \begin{tabular}{@{}llp{0.19\textwidth}p{0.17\textwidth}lcp{0.21\textwidth}@{}}
 \toprule
 Item & Project & Template (tier) & Checkers & Rater & Conf. & Reason \\
 \midrule
 A01 & fantasy\_hotel\_2 & compound of three edits (1) & fail: intent (slab elevation off by 0.1 m) & cannot judge & low & elevations after the edit not shown \\
 A02 & city\_house\_munich & set wall height, spatial reference (1) & pass & yes & high & \\
 A03 & fantasy\_hotel\_2 & delete slab and its containment (1) & fail: validity, preservation (5,314 side effects) & no & high & side effects, dependents, invalid: 679 elements changed \\
 A04 & fantasy\_office\_building\_1 & set wall height (1) & pass & yes & medium & \\
 A05 & smiley\_west & create door, path reference (1) & pass & no & medium & door placed in the wall on the other side of the route \\
 A06 & molio & re-host window (1) & fail: intent (file unchanged) & no & high & requested change missing \\
 A07 & duplex & move wall with openings (1) & fail: intent, dependency (nothing moved) & no & high & requested change missing \\
 A08 & fantasy\_hotel\_1 & compound of two edits (1) & fail: intent (column at the wrong position) & no & high & requested change wrong \\
 A09 & fantasy\_office\_building\_3 & compound of three edits (1) & pass & cannot judge & low & slab thickness and top not shown; view occluded \\
 A10 & molio & delete column, spatial reference (1) & pass & cannot judge & low & reference wall not identifiable in the view \\
 A11 & city\_house\_munich & widen opening (1) & fail: intent (width 1.00 m instead of 1.20 m) & cannot judge & low & widths not shown \\
 A12 & duplex & create slab under a room (1) & pass & cannot judge & low & thickness and elevation not shown \\
 A13 & dental\_clinic & move wall bounding spaces (2) & pass & yes & medium & \\
 A14 & smiley\_west & move joined wall (3) & fail: intent, dependency (unchanged) & no & high & requested change missing \\
 A15 & schependomlaan & delete column (1) & pass & yes & high & \\
 A16 & molio & create wall (1) & pass & yes & medium & \\
 A17 & digital\_hub & create slab on a space footprint (1) & fail: intent (footprint in local instead of world coordinates) & no & high & slab near the origin, not under the space \\
 A18 & fantasy\_hotel\_2 & create slab on columns (1) & fail: intent (footprint) & cannot judge & low & slab outside the section box; no dimensions \\
 A19 & molio & move column towards the exterior wall (1) & pass & cannot judge & low & move confirmed, nearest exterior wall not identifiable \\
 A20 & smiley\_west & delete wall, viewpoint reference (1) & fail: intent, dependency, preservation (wrong wall) & no & medium & requested change wrong, side effects \\
 A21 & samuel\_macalister & delete the lowest slab, spatial reference (1) & fail: intent, preservation (wrong slab) & cannot judge & low & elevations not shown \\
 A22 & city\_house\_munich & move joined wall (3) & fail: preservation (one unlisted space modified) & yes & medium & wall, window, joined walls and adjoining space all consistent \\
 A23 & fantasy\_hotel\_2 & resize slab (1) & pass & yes & high & \\
 A24 & schependomlaan & compound of two edits (1) & fail: intent, dependency (no connection relationship) & no & high & requested change wrong, dependents \\
 A25 & smiley\_west & move wall bounding spaces (2) & fail: intent, dependency (unchanged) & no & high & requested change missing \\
 A26 & fantasy\_hotel\_1 & set space height (1) & pass & yes & medium & \\
 A27 & 4351 & move wall with openings (1) & fail: intent, dependency (unchanged) & no & high & requested change missing \\
 A28 & smiley\_west & re-host window (1) & pass & yes & medium & \\
 A29 & city\_house\_munich & move joined wall (3) & pass & cannot judge & low & opening change outside the pictured area \\
 A30 & molio & delete space (1) & pass & yes & high & \\
 \bottomrule
 \end{tabular}
\end{table}

\paragraph{Results on the 20 pairs}
Table~\ref{tab:supp_he_pairs} lists every pair with the benchmark scores of both outputs and the rater's choice. One pair was left unanswered because the benchmark instruction contains a malformed coordinate; of the 19 answered pairs the final student was preferred in 10, tied as acceptable in 7, and the other output preferred in 2. Where a side was preferred, the choice agrees with the sign of the benchmark score in 10 of 12 pairs; in the two others the rater preferred the output that made the requested change in full (a window raised together with its opening, pair B34) or made a change at all (pair B50, where the released GPT-5.4 Pro output leaves the model unchanged but carries the higher official score). In two BIM-Edit pairs against GPT-5.4 the rater found both outputs unacceptable and still expressed a preference for the student's.

\begin{table}[!htbp]
 \centering
 \scriptsize
 \setlength{\tabcolsep}{3pt}
 \renewcommand{\arraystretch}{1.08}
 \caption{The 20 benchmark pairs: task, the other system, the benchmark score of the final student and of the other output (BIM-Edit: our scorer for our outputs and the official score for GPT-5.4 Pro; BIBIMBAP: the benchmark's score), the rater's preference and the acceptability of each output (final student / other).}
 \label{tab:supp_he_pairs}
 \begin{tabular}{@{}llp{0.25\textwidth}lccp{0.22\textwidth}@{}}
 \toprule
 Pair & Benchmark & Task & Other & Score student / other & Preference & Acceptable (student / other) and note \\
 \midrule
 B31 & BIM-Edit & create door between two rooms & GPT-5.4 Pro & 0.65 / 0.00 & student & no / no; door overshoots the wall end \\
 B32 & BIBIMBAP & rotate wall by a quarter turn & 9B SFT & 0.67 / 0.33 & tie & yes / yes; different pivots, both legitimate \\
 B33 & BIBIMBAP & delete columns unrelated to a slab & 9B SFT & 0.50 / 0.00 & student & yes / unsure \\
 B34 & BIBIMBAP & raise window by 0.5 m & 27B SFT & 0.33 / 0.00 & 27B SFT & no / yes; student raised the opening only \\
 B35 & BIM-Edit & delete wall, spatial reference & 9B SFT & 0.58 / 0.58 & tie & yes / yes; identical results \\
 B36 & BIBIMBAP & move column by 3 m & 9B SFT & 1.00 / 1.00 & tie & yes / yes \\
 B37 & BIBIMBAP & remove the wall touching two walls & 27B SFT & 0.50 / 0.00 & tie & yes / yes; two walls satisfy the condition \\
 B38 & BIBIMBAP & delete opening & 27B SFT & 0.50 / 0.00 & student & yes / no; other made no change \\
 B39 & BIM-Edit & delete room, topological reference & GPT-5.4 Pro & 0.67 / 0.67 & tie & yes / yes \\
 B40 & BIBIMBAP & enclose a room & 9B SFT & 1.00 / 0.00 & student & yes / no \\
 B41 & BIBIMBAP & rotate wall by a quarter turn & 27B SFT & 0.67 / 1.00 & tie & yes / yes \\
 B42 & BIBIMBAP & rotate parallel wall by 45 degrees & 27B SFT & 1.00 / 1.00 & tie & yes / yes \\
 B43 & BIBIMBAP & rotate parallel wall by 45 degrees & 9B SFT & 1.00 / 0.67 & student & yes / no; other unchanged \\
 B44 & BIM-Edit & create wall, topological reference & 9B SFT & 0.37 / 0.00 & student & yes / no \\
 B45 & BIM-Edit & create slab at given coordinates & 9B SFT & 0.84 / 0.30 & unanswered & malformed coordinate in the instruction \\
 B46 & BIM-Edit & delete wall, spatial reference & 9B SFT & 0.67 / 0.36 & student & yes / no \\
 B47 & BIM-Edit & update window, topological reference & GPT-5.4 Pro & 0.55 / 0.00 & student & no / no; window kept in the wrong storey \\
 B48 & BIM-Edit & create wall with four connections & 9B SFT & 0.00 / 0.47 & 9B SFT & no / no; other partial, student unchanged \\
 B49 & BIM-Edit & create window, spatial reference & GPT-5.4 Pro & 0.58 / 0.00 & student & unsure / no \\
 B50 & BIM-Edit & update wall, spatial reference & GPT-5.4 Pro & 0.42 / 0.81 & student & no / yes; GPT-5.4 output unchanged \\
 \bottomrule
 \end{tabular}
\end{table}

\clearpage
\begingroup
\setlength{\intextsep}{2pt}
\setlength{\abovecaptionskip}{3pt}
\section{Further visual examples}
\label{sec:supp_examples}

Figure~\ref{fig:supp_ifceditbench} shows the before-and-after section views used in the expert evaluation. The four passing examples are: a door created in a wall that the instruction names by a route through the building (a), a window re-hosted to the parallel wall on the other side of its room (b), a wall move that drags the footprints of the spaces it bounds (c) and a wall move at the deepest tier, where the joined walls, their openings and the spaces all follow (d). The four failures show what each checker rejects: a slab created with the wrong footprint fails intent (e); a wall move that the student did not carry out (eleven inspection calls, no commit) fails intent and dependency (f); a wall move that also alters a space outside the allowed set fails preservation (g); and a slab deletion whose removal cascaded through geometry shared with other products fails validity as well as preservation (h). Section~\ref{sec:supp_case_audit} audits the relationships and attributes behind four of these verdicts.

\begin{figure}[!htbp]
 \centering
 \includegraphics[width=\textwidth]{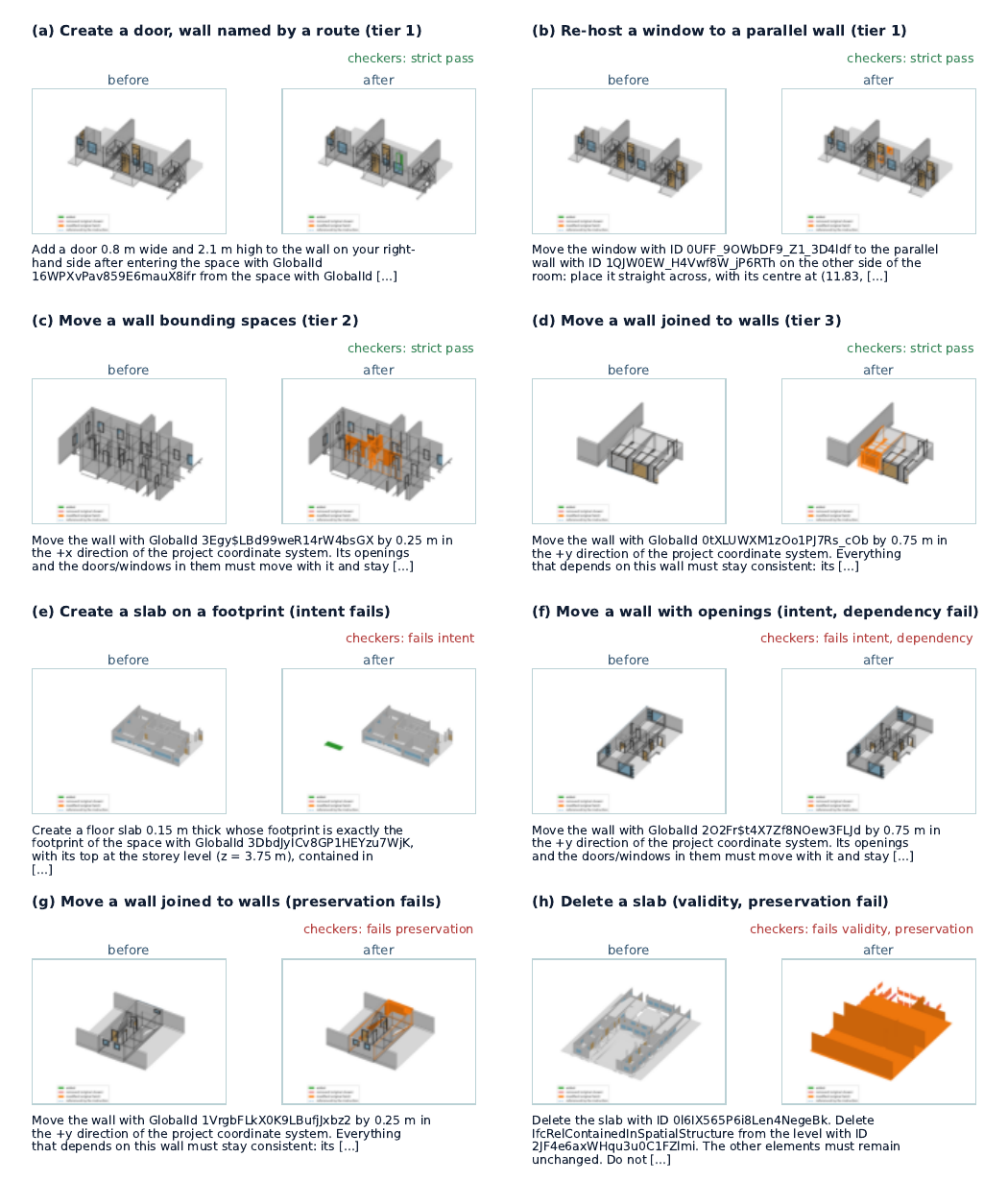}
 \caption{IfcEditBench edits of the final student: before and after each edit (green = added, red = removed, orange = modified, dashed = elements the instruction refers to) with the instruction and the four-gate verdict. (a)--(d) strict passes at tiers 1 to 3; (e)--(h) checker failures with the failing gates.}
 \label{fig:supp_ifceditbench}
\end{figure}

\subsection{Relationship and attribute audit of four cases}
\label{sec:supp_case_audit}
The entity and relationship diffs explain the checker verdicts in Figure~\ref{fig:supp_ifceditbench}. (b) Window re-hosted: two products modified, the window and its opening; one relationship removed and one added, the opening's voids relationship moving from the old wall to the new one; the intent test finds the opening voiding exactly the new host, its plan centre within 0.000 m of the requested point and its sill at the requested height. (c) and (d) Wall moves at tiers two and three: the wall's centroid moved by the requested vector within 0.000 m; the opening and the fill translated with it through their relative placement and the opening still voids the wall; at tier two the three bounded spaces' faces and their space-boundary curves lie on the moved wall within 0.02 m, as they did before; at tier three the surface gap between each joined wall and the moved wall is no larger after the move than before; no product outside the allowed set changed. (g) Tier-three move rejected: the same tests pass, but the diff lists one modified product outside the allowed set, a basement space whose footprint the student shifted by the move vector because it abutted the wall face; the Revit 2025 export carries no space boundaries, so the generator had no relationship from which to list that space as a dependent. (h) Slab deletion rejected: the instruction asks to delete a footing slab and to remove it from the storey's containment relationship, that is, to drop one member of the relationship instance rather than the instance itself. The student's commit removed the slab, but the deletion cascaded through geometry entities that this millimetre Revit export shares between products: 273 products removed, 406 modified, 4,638 relationship instances dropped (openings and containment on other storeys among them; 5,314 of these changes lie outside the allowed set) and 674 new schema errors from placements and faces left without their referents. Although the student's final message says that only the slab was removed and the relationship kept, the diff records damage to shared geometry and unrelated products. The rater rejected the edit for these side effects.

\clearpage

Figure~\ref{fig:supp_benchpairs} shows six public-benchmark tasks with the final student's output next to another system's output. On BIM-Edit, the student places a door in the wall between two named rooms where the released GPT-5.4 Pro output scores 0.00 (a), deletes a room with the same partial score as GPT-5.4 Pro (b) and, like its 9B SFT start, deletes the wall on the left after walking into a room with a partial score (c). These partial scores reflect BIM-Edit's separate measures of geometry, semantics and topology. On BIBIMBAP, the final student rotates a wall by a quarter turn (d) and deletes the columns unrelated to a slab (e) where the SFT start scores 0.33 and 0.00, and in (f) the final student raises only the opening of a window by half a metre, scored 0.33, where the 27B SFT student raises window and opening together but scores 0.00.

\begin{figure}[!htbp]
 \centering
 \includegraphics[width=\textwidth]{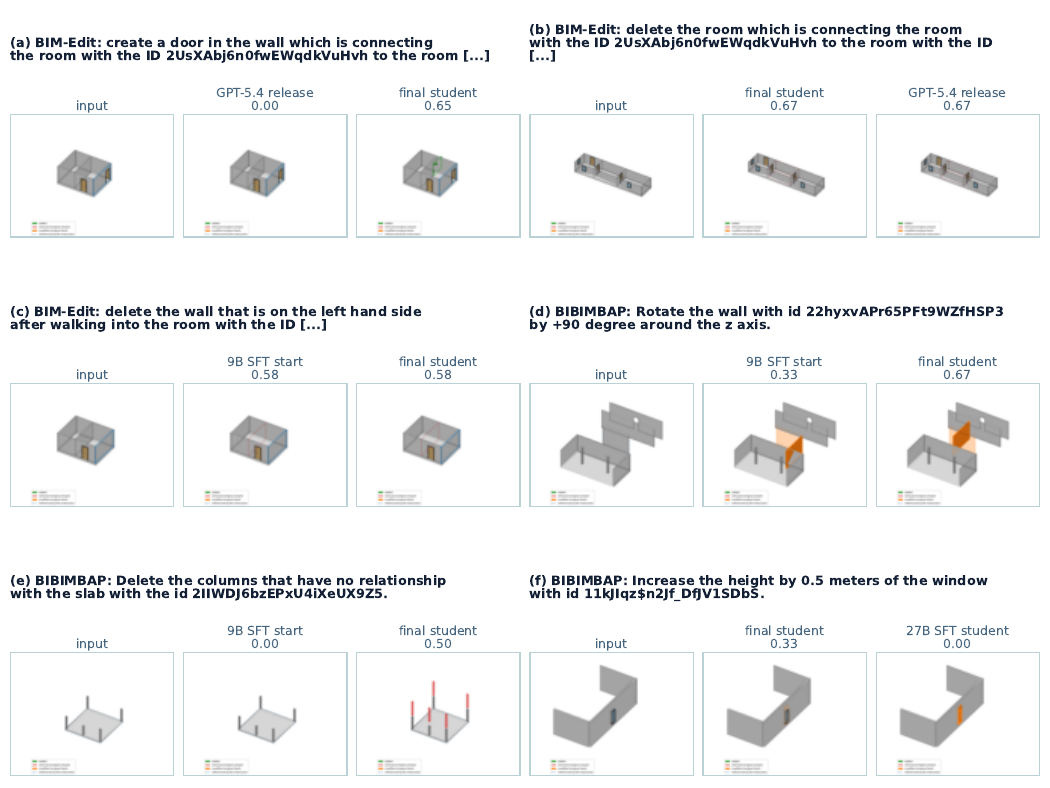}
 \caption{Public-benchmark tasks: the input model, the final student's output and another system's output (the released GPT-5.4 Pro outputs, the 9B SFT start or the 27B SFT student), with the benchmark score of each output where the benchmark defines one.}
 \label{fig:supp_benchpairs}
\end{figure}

\endgroup

\end{document}